\documentclass{article}

\usepackage{iclr2027_conference,times}

\usepackage{amsmath,amsfonts,bm}

\def\eqref#1{equation~\ref{#1}}

\def\1{\bm{1}}

\DeclareMathAlphabet{\mathsfit}{\encodingdefault}{\sfdefault}{m}{sl}
\SetMathAlphabet{\mathsfit}{bold}{\encodingdefault}{\sfdefault}{bx}{n}

\usepackage{amsmath,amssymb}
\usepackage{booktabs}
\usepackage{multirow,array,colortbl}
\usepackage{needspace,flafter}
\usepackage{wrapfig,caption}
\usepackage{placeins,float}
\definecolor{DraftBlue}{RGB}{42,83,158}
\definecolor{MethodTint}{RGB}{236,244,248}
\usepackage{graphicx}
\usepackage{hyperref}
\usepackage{url}

\hypersetup{hidelinks}

\title{AlphaDiverse: Post-Training Local\\
Quantitative Research Agents for Diverse\\
Exploration in Alpha Factor Mining}

\author{\parbox{\dimexpr\textwidth-2\tabcolsep\relax}{
Qingzhuo Wang$^{1,2}$\thanks{Work done during an internship at Shanghai Non-convex Intelligent Technology.}\quad
Zikun Wei$^{2}$\quad
Zhihua Wei$^{1}$\quad
Wen Shen$^{1}$\thanks{Corresponding Author}\\[4pt]
\normalfont $^{1}$Tongji University\quad
$^{2}$Shanghai Non-convex Intelligent Technology}}

\iclrfinalcopy  

\begin{document}

\maketitle

\begin{abstract}
Large language model (LLM)-based multi-agent systems can automate alpha factor mining, but their reliance on external APIs limits control over cost, availability, and confidentiality. Long research loops also tend to revisit a few successful economic mechanisms that lead to research path collapse. To address these limitations, we propose \textsc{AlphaDiverse}, a framework that integrates a multi-agent alpha research system, diverse research path collection, and post-training for local agents. We let the research system generate complementary plan portfolios and vary research environments across loops to collect diverse research paths. Using these diverse traces, we warm-start local Planner and Realizer agents with supervised fine-tuning. Then, we propose a joint GRPO method to optimize both of them using predictive quality and diversity of contributions. Research feedback is confined to inner period data, while a frozen final model is evaluated on a later outer period data, thereby avoiding test-set tuning. Experiments across four Chinese stock universes show that \textsc{AlphaDiverse} can combine competitive prediction with broader exploration.

\end{abstract}

\section{Introduction}
\label{sec:introduction}

Alpha factor mining is a central task in quantitative research, aiming to construct signals from market data that help predict future returns. Machine learning and deep learning have advanced this task through predictive representation learning, automated factor construction, and the combination of signals from high-dimensional market data \citep{gu2020empirical,yang2020qlib,duan2022factorvae,chen2024deeplearning,li2024master}. Recently, many studies have used LLMs to empower quantitative research \citep{yang2023fingpt,xiao2024tradingagents,xiong2025flagtrader,zhang2026quantevolver}. For alpha factor mining, especially, multi-agent systems organize hypothesis generation, factor implementation, backtesting, and iterative refinement across specialized agents \citep{li2025rdagent,tang2025alphaagent,guo2026aqua}. Although advanced multi-agent systems can cover more of the research process, current systems rely primarily on external APIs and still face two practical limitations: (1) repeated calls to external APIs are costly, and it is hard to control the APIs' availability, latency, and model behavior; and (2) proprietary data sources and research paths often require strict confidentiality, which external APIs cannot guarantee.

Furthermore, long-horizon agent loops and auto research systems can suffer from \textit{research path collapse}. Systems may repeatedly extend a few successful mechanisms proposed early and explore only a small part of the large search space \citep{tang2025alphaagent,shi2026alphajungle,audranreiss2025ideation,chen2026diversitycollapse,liu2026cognitive}. We observe a similar pattern in our alpha factor mining experiments: after one hypothesis family yields a good result, later rounds often revisit the same economic mechanism, leaving much of the search space unexplored. Formula- and code-level checks can detect duplicate expressions or implementations \citep{tang2025alphaagent,hubble,zhang2026quantevolver,liu2026cognitive}, but they do not by themselves prevent repeated exploration of the same economic mechanism.

\begin{figure}[t]
  \centering
  \includegraphics[width=\linewidth]{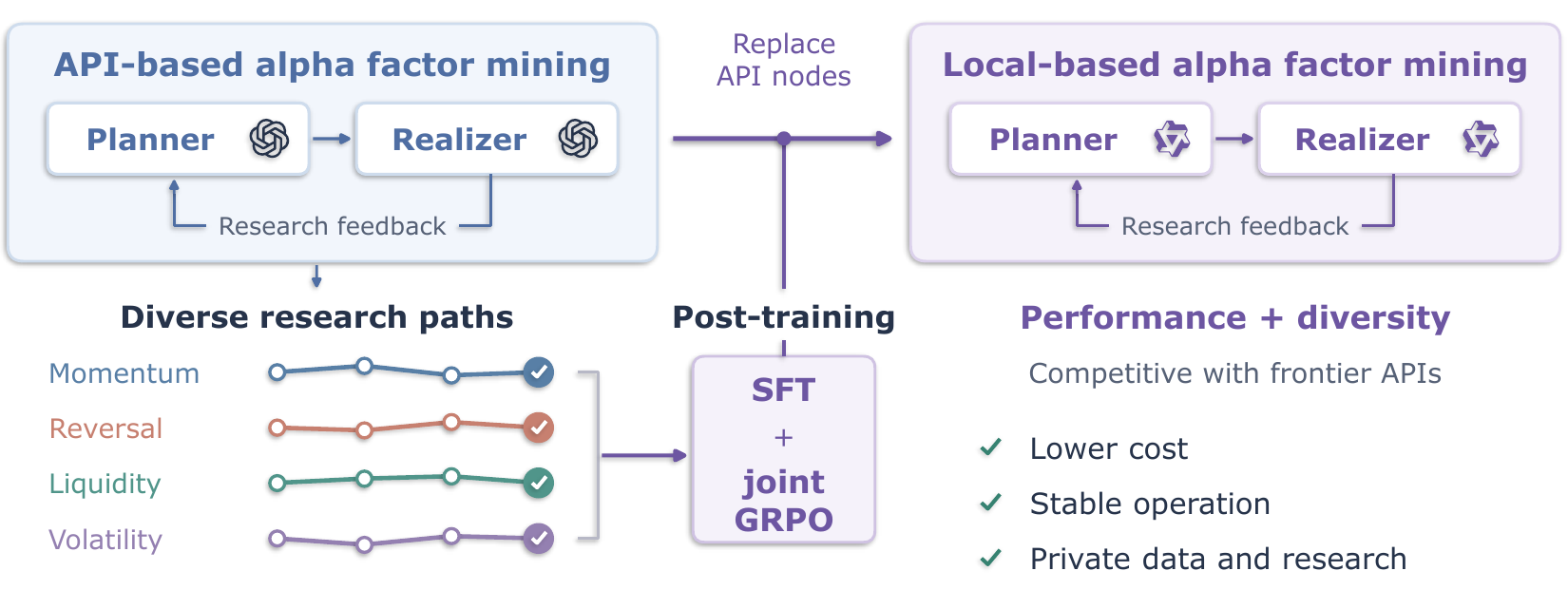}
  \caption{Illustration of the proposed \textsc{AlphaDiverse} framework. API-based research loops produce diverse plans, factor implementations, and evaluation feedback. We select high-quality, complementary traces to warm-start local Planner and Realizer agents, then jointly refine them with GRPO. The trained nodes replace their API counterparts for locally controlled alpha factor research.}
  \label{fig:overview}
\end{figure}

To address these limitations, we introduce \textsc{AlphaDiverse}, a framework that integrates multi-agent alpha factor research, diverse research path collection, and  post-training of local agents. During the alpha factor research, \textsc{AlphaDiverse} iteratively proposes research plans, realizes factors, and evaluates their predictive quality. As Figure~\ref{fig:overview} shows, we first use advanced closed-source LLMs such as GPT through APIs to verify that the loop can complete this workflow reliably. The planner agent proposes research plans, and the realizer agent turns them into executable factors. In this way, we obtain diverse research paths covering various economic mechanisms from these loops, such as momentum, reversal, liquidity, volatility, etc. Using these paths, we are, to our knowledge, the first to post-train local agents based on models such as Qwen for both research planning and factor realization within a multi-agent alpha factor mining workflow, replacing the API nodes with locally deployed agents.
This removes their external API calls and keeps proprietary data and research trajectories inside the organization, while making their cost, availability, and behavior stable.

Furthermore, to broaden exploration beyond a few successful research paths, we introduce a two-level exploration strategy: within each round of each alpha factor mining loop, the planner proposes a set of plans grounded in complementary economic mechanisms; across loops, we vary stock markets and available features to collect diverse research paths. We then select high-quality, complementary traces and balance their representation as training traces. Based on this, we apply two-stage post-training as Figure~\ref{fig:overview} shows: At stage one, SFT teaches instruction-following capability and quantitative research knowledge. At stage two, we propose a joint GRPO method to optimize both agents using shared rewards for predictive quality and complementary factor contributions, with credit assigned separately to planning and realization. This design improves both exploration diversity and generalization beyond the paths observed during SFT.

Finally, repeated feedback in existing alpha-mining systems~\citep{li2025rdagent,tang2025alphaagent,han2026quantaalpha} can lead to \textit{test-set tuning}, where the same data guide iterative research and are later used to report final performance~\citep{luo2025pitfalls,ning2026molprop,ning2026materials,li2026autoscientistquant}. We divide the data into an inner period for research and an outer period for final evaluation. Agents receive feedback only from the inner period during the research loop. For the final evaluation, the selected factors and model are frozen, thereby the final model is only evaluated once on the outer period thus avoid test-set tuning. Within the inner period, we construct multiple chronological folds, each with its own training, validation, and test segments, and aggregate their results to guide the loop. We find that varying the number of folds also produces different research paths and training traces. We evaluate our method across four Chinese stock universes. The results show consistent improvements in both research performance and the diversity of explored research paths after local post-training.

\section{Methodology}
\label{sec:methodology}

\begin{figure}[t]
  \centering
  \includegraphics[width=\linewidth]{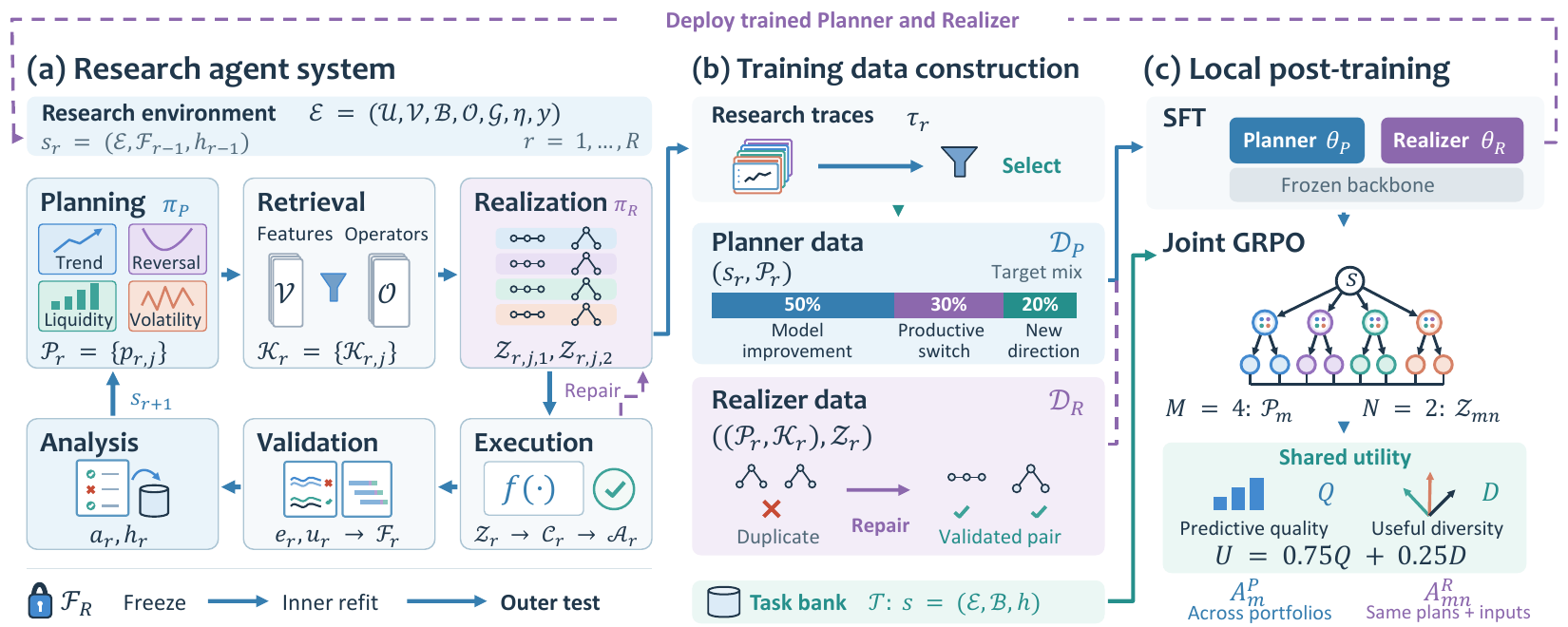}
  \caption{\textsc{AlphaDiverse} overview. (a) The research loop proposes plans, implements factors, and updates its state from inner evaluation. (b) These records supply selected Planner demonstrations, Realizer examples, and RL tasks. (c) SFT and joint GRPO train two local policies, which replace the corresponding research nodes.}
  \label{fig:architecture}
\end{figure}

As Figure~\ref{fig:architecture} shows, \textsc{AlphaDiverse} connects a research agent system, diverse training traces construction, and local post-training.
Given a research environment $\mathcal E=(\mathcal U,\mathcal V,\mathcal B,\mathcal O,\mathcal G,\eta,y)$, each research loop aims to mine a set of alpha factors $\mathcal F$ that transform historical observations into signals predictive of future returns. Here $\mathcal U$ is the stock universe, $\mathcal V$ is the available market features, $\mathcal B$ is the initial factor set, $\mathcal O$ is the permitted operators and input transforms, $\mathcal G$ is the feasible economic-mechanism catalogue, $\eta$ denotes the evaluation settings, and $y$ is the return target. The catalogue $\mathcal G$ contains mechanisms supported by $(\mathcal V,\mathcal O)$, grouped by economic theme. Please see Appendix~\ref{app:method-examples} for the example of a research environment. When mining factors, only observations available at prediction time can be used; a downstream predictor combines them to predict the return target $y$. Each loop has $R$ rounds. Its round-$r$ state is denoted by $s_r=(\mathcal E,\mathcal F_{r-1},h_{r-1})$, where $h_{r-1}$ summarizes earlier plans, implementations, and feedback, and $\mathcal F_0=\mathcal B$.

Given the state $s_r$ of round $r$, the Planner proposes $J$ plans $\mathcal P_r=\{p_{r,j}\}_{j=1}^{J}$, and the Realizer produces their factor specifications $\mathcal Z_r$. A research path is the sequence of mechanisms and implementations explored over these rounds. We obtain diverse research paths by varying $\mathcal E$. We study local policies that improve the predictive value of $\mathcal F_R$ while exploring complementary mechanisms under this fixed budget $R$. 

\subsection{Research Agent System}
\label{sec:agent-system}

The research agent system contains six modules, including planning, retrieval, realization, execution, validation and analysis, organized into a research loop.

\textbf{Planning.}
Each plan $p_{r,j}$ generated by the Planner $\pi_{\mathrm P}$ is based on a mechanism in $\mathcal G$ and contains an event, economic explanation, horizon, condition, expected direction, and relation to existing factors. Please see  Appendix~\ref{app:method-examples} for an example of a plan. We use $J$ distinct mechanisms covering at least three themes:
\begin{small}
\begin{equation}
  \mathcal P_r=\{p_{r,j}\}_{j=1}^{J}\sim\pi_{\mathrm P}(\cdot\mid s_r),
  \qquad
  s_r=(\mathcal E,\mathcal F_{r-1},h_{r-1}).
  \label{eq:planning}
\end{equation}
\end{small}

\textbf{Capability retrieval and realization.}
Given a portfolio of plans $\mathcal P_r$, the retrieval module selects a set of features from $\mathcal V$ and operators from $\mathcal O$ for each plan, denoted by $\mathcal K_r=\{\mathcal K_{r,j}\}_{j=1}^{J}$. The Realizer $\pi_{\mathrm R}$ then generates two different factor specifications $\mathcal Z_{r,j}=(\mathcal Z_{r,j,1}, \mathcal Z_{r,j,2})$ per plan:
\begin{small}
\begin{equation}
  \mathcal K_{r,j}=\operatorname{Retrieve}(p_{r,j};\mathcal V,\mathcal O),
  \qquad
  \mathcal Z_r\sim\pi_{\mathrm R}(\cdot\mid\mathcal P_r,\mathcal K_r).
  \label{eq:plan-realization}
\end{equation}
\end{small}
Each factor specification declares the operator, inputs, transforms, required parameters, direction, and rationale. We limit $\mathcal K_{r,j}$ to be a relatively small set, in order to reduce implementation ambiguity and simplify context. We restrict the two factor specifications $\mathcal Z_{r,j,1}$ and $\mathcal Z_{r,j,2}$ to be structurally different in order to provide alternative realizations of the same mechanism. They must differ in their computational structure, not merely in their name or window. Appendix~\ref{app:method-examples} illustrates these interfaces.

\textbf{Execution and factor screening.}
The execution module checks the validity of factor specifications $\mathcal Z_r$ and compiles valid specifications into executable factor formulas. A specification for plan $p_{r,j}$ is valid only if it uses the retrieved features and operators $\mathcal K_{r,j}$, satisfies their input and parameter constraints, and uses only information available at prediction time. Otherwise, the module will reject the specification because it uses invalid features or operators, or encounters future information, resulting in lookahead leakage. Invalid specifications will be fed back to the Realizer $\pi_{\mathrm R}$ with a repair notice. Executing valid specifications produces candidate factors $\mathcal C_r$, which are screened for coverage, predictive quality, and redundancy under the evaluation settings $\eta$:
\begin{small}
\begin{equation}
  \mathcal C_r=\operatorname{Execute}(\mathcal Z_r;\mathcal K_r),
  \qquad
  \mathcal A_r=\operatorname{Screen}(\mathcal C_r;y,\mathcal F_{r-1},\eta).
  \label{eq:factor-screening}
\end{equation}
\end{small}
For coverage, we measure the factor's effective coverage across all trading days. For predictive quality, we evaluate the factor's predictive performance relative to the return target $y$. For redundancy, we assess the factor's correlation with the current factors $\mathcal F_{r-1}$ and other factors within $\mathcal C_r$. Each of these three metrics has a corresponding acceptance threshold. A factor is added to $\mathcal A_r$ only if it meets all the criteria. Please see Appendix~\ref{app:factor-screening-details} for details on the metrics and threshold settings.

\textbf{Predictor validation and factor retention.}
The evaluation settings $\eta$ chronologically divide data into an inner period and an outer period, we only use inner data in the loop to avoid \textit{test-set tuning}. The inner period data is divided into $K$ folds, each fold contains chronological training, validation, and inner test segments. The inner folds serve as nonoverlapping search windows: validation selects predictor settings, and all results of the windows supply loop feedback. Varying $K$ changes this evidence and can \textit{produce different paths}. We choose LightGBM as the predictor.

To determine whether $\mathcal A_r$ should be added to $\mathcal F_r$ and whether the best predictor should be updated, we evaluate $\mathcal F_{r-1}\cup\mathcal A_r$ against the previous best factor set $\mathcal F_{r-1}$ by a prediction score. Let $S_k(\mathcal F)$ denote the prediction score of factor set $\mathcal F$ on inner fold $k$. The score is computed from predictions produced by a predictor using $\mathcal F$ as input. (Definition of the score appears in Appendix~\ref{app:factor-screening-details}). The retention decision $u_r\in\{0,1\}$ is:
\begin{small}
\begin{equation}
  u_r=\mathbf 1\!\left[
    \sum_{k=1}^{K}w_k\Delta_{r,k}>0
    \ \land\
    \sum_{k=1}^{K}\mathbf 1[\Delta_{r,k}>0]\geq m_K
  \right],\quad s.t.\ \Delta_{r,k}=S_k(\mathcal F_{r-1}\cup\mathcal A_r)-S_k(\mathcal F_{r-1})
  \label{eq:model-retention}
\end{equation}
\end{small}
where the normalized weights $w_k$ are proportional to the number of search days, and $m_K$ requires a majority of folds to improve (e.g. two of three). Equation~\ref{eq:model-retention} sets $u_r=1$ only when adding $\mathcal A_r$ improves the weighted model score across the $K$ inner folds and increases the score on at least $m_K$ folds. If $u_r=1$, we set $\mathcal F_r=\mathcal F_{r-1}\cup\mathcal A_r$ and update the settings of the best predictor.

\textbf{Analysis and the next state.}
Let $e_r$ represent the evaluation evidence with factors and predictor. The analysis module maps $(\mathcal P_r,\mathcal Z_r,e_r,u_r)$ to feedback $a_r$ analyzing the result of the current round and suggesting the next research step. Updating $h_{r-1}$ with these records gives $h_r$ and hence $s_{r+1}=(\mathcal E,\mathcal F_r,h_r)$.

\textbf{Final evaluation.}
After $R$ rounds, we freeze $\mathcal F_R$ and the settings of the predictor, then refit the predictor within inner data, and evaluate the final predictor once on outer data. 

\subsection{Training Data Construction}
\label{sec:training-data}

Each round in Section~\ref{sec:agent-system} yields a record $\tau_r=(s_r,\mathcal P_r,\mathcal K_r,\mathcal Z_r,e_r,u_r,a_r)$. Collecting these records across environments provides two supervision interfaces: $(s_r,\mathcal P_r)$ for the Planner dataset $\mathcal D_{\mathrm P}$ and $((\mathcal P_r,\mathcal K_r),\mathcal Z_r)$ for the Realizer dataset $\mathcal D_{\mathrm R}$. 

\textbf{Planner data construction.}
To construct suitable training data for the Planner, we first identify appropriate state--portfolio pair samples $(s_r, \mathcal{P}_r)$, selecting only those where at least two plans include an accepted factor (see the Execution and factor screening paragraph in Section~\ref{sec:agent-system}). We then categorize these samples into three groups: (1) Model improvement sample, where $u_r=1$ in the current round. These samples are used to improve the quality of the Planner's research; (2) Productive switch samples, where $u_{r-1}=0$  and at least half of the factors were rejected in round $r-1$, and at least one plan with accepted factors in the current round uses a mechanism different from round $r-1$. These samples are used to enhance the Planner's ability to switch direction after failure; and (3) New direction samples, where at least one plan uses a mechanism never used in prior rounds and the corresponding factor is accepted. These samples are used to improve the Planner's ability to explore new directions. We target a 50/30/20 mixture of these groups, balancing environments, API models, and research stages within each group and limiting repeated mechanisms. Appendix~\ref{app:selection-rules} gives the assignment order and thresholds.

\textbf{Realizer data construction.}
We also construct appropriate training data for the Realizer by selecting suitable pair samples $((\mathcal P_r,\mathcal K_r),\mathcal Z_r)$. We require the two generated factor specifications $\mathcal Z_{r,j,1}$ and $\mathcal Z_{r,j,2}$ per plan $\mathcal P_{r,j}$ to meet the quality and correlation thresholds (Please see Appendix~\ref{app:realizer-data-details} for details). These examples teach different implementations under identical research intent and available inputs. When the factor specification generated by the Realizer is invalid, the system requires Realizer to repair it. Thus, we also need to construct samples for this repair process. We create repair samples by duplicating a specification, changing only its window or repeating an input, etc. The supervision target remains the original, validated pair. This data construction provides supervision for distinct implementations and facilitates the correction of the realization process.

\subsection{Local Post-Training}
\label{sec:local-post-training}

\textbf{Supervised warm start.}
We first use SFT to teach the local Planner to propose economic research plans and the local Realizer to implement them as valid factor specifications. The two policies share a frozen pretrained backbone and use separate LoRA adapters~\citep{hu2021lora}, with parameters $\theta_{\mathrm P}$ and $\theta_{\mathrm R}$. Section~\ref{sec:training-data} provides input--response pairs $(c,v)$ for each role $a\in\{\mathrm {Planner},\mathrm {Realizer}\}$. For the Planner, $c=s_r$ and $v=\mathcal P_r$; for the Realizer, $c=(\mathcal P_r,\mathcal K_r)$ and $v=\mathcal Z_r$. We train each policy by minimizing
\begin{small}
\begin{equation}
  \mathcal L_{\mathrm{SFT}}(\theta_a)
  =-\frac{1}{N_a}\sum_{(c,v)\in\mathcal D_a}\sum_{\ell=1}^{|v|}
    \log\pi_{\theta_a}(v_\ell\mid c,v_{<\ell}),
  \qquad N_a=\sum_{(c,v)\in\mathcal D_a}|v|.
  \label{eq:role-sft}
\end{equation}
\end{small}
Here $v_\ell$ is the $\ell$-th response token, and $N_a$ is the total number of response tokens for role $a$. For repair examples, the input also contains the invalid response and its validation feedback. Only the final assistant response is supervised. This warm start supplies the instruction-following ability and research knowledge used in subsequent joint reinforcement learning. 

\textbf{Joint exploration from research states.}
Starting from the SFT policies, we jointly optimize the Planner and Realizer with GRPO~\citep{shao2024deepseekmath}. We construct a task bank $\mathcal T$ from the environments and histories collected in Section~\ref{sec:training-data}. Each state $s=(\mathcal E,h)$ specifies an environment and earlier research summarized in $h$. For the same state $s$, we sample $M$ plan portfolios and $N$ independent Realizer responses for each portfolio
\begin{small}
\begin{equation}
  \mathcal P_m\sim\pi_{\theta_{\mathrm P}}(\cdot\mid s),
  \qquad
  \mathcal Z_{mn}\sim\pi_{\theta_{\mathrm R}}(\cdot\mid\mathcal P_m,\mathcal K_m),
  \quad m=1,\ldots,M,\ n=1,\ldots,N,
  \label{eq:joint-sampling}
\end{equation}
\end{small}
where $\mathcal K_m$ is retrieved for $\mathcal P_m$ as in Equation~\ref{eq:plan-realization}. Thus, the $N$ responses for one portfolio are alternative implementations of the same plans and retrieved content. Execution and screening evaluate each response separately, yielding accepted factors $\mathcal A_{mn}$ and per-plan subsets $\mathcal A_{mn,j}$. A LightGBM evaluator then measures their contribution on a later inner reward window. Please see Appendix~\ref{app:joint-task-bank} for task construction, nested sampling, and the LightGBM evaluation protocol.

\textbf{Rewarding predictive quality and useful diversity.}
We reward both improvements in prediction and useful contributions beyond earlier research. Consider one sampled portfolio $\mathcal P$ with accepted factors $\mathcal A$; we omit the sampling indices $m,n$ below. Let $S_s(\mathcal F)$ be the reward-window score of the LightGBM model using factor set $\mathcal F$. It uses the same model score defined in Appendix~\ref{app:factor-screening-details}. The score gain is $\Delta_s(\mathcal A)=S_s(\mathcal B\cup\mathcal A)-S_s(\mathcal B)$, where $\mathcal B\in\mathcal E$ is the baseline factor set. We define the portfolio quality $Q$ and the quality $q_j$ of each plan evaluated separately against the baseline as
\begin{small}
\begin{equation}
  Q=\tanh\!\left(\frac{\Delta_s(\mathcal A)}{\sigma_s}\right),
  \qquad
  q_j=\operatorname{clip}\!\left(
     \frac{\Delta_s(\mathcal A_j)-\kappa_s}{\sigma_s},0,1\right),
  \label{eq:reward-quality}
\end{equation}
\end{small}
where $\sigma_s>0$ scales the score gains and $\kappa_s\geq0$ sets the minimum gain for a plan to receive positive quality weight. Please see Appendix~\ref{app:noise-controls} for the detailed definition.

For stock $i$ on reward date $t$, adding plan $j$ changes the baseline prediction by $d_{j,i,t}=\widehat y^{(j)}_{i,t}-\widehat y^{(0)}_{i,t}$. We subtract the daily mean $\bar d_{j,t}=(\sum\nolimits_i d_{j,i,t})/n_t$, concatenate the weighted differences as $\widetilde b_j=\operatorname{concat}_{i,t}[(d_{j,i,t}-\bar d_{j,t})/\sqrt{T_s n_t}]$, and normalize $b_j=\widetilde b_j/\|\widetilde b_j\|_2$. Here $n_t$ counts eligible stocks on date $t$, and $T_s$ counts reward dates. Similar $b_j$ indicate similar prediction changes. Please see Appendix~\ref{app:prediction-direction-details} for detailed definitions.

We evaluate earlier plans summarized in $h$ in the same way, obtaining an archive $\mathcal H_s$ of qualities and prediction-change vectors. For any collection of evaluated plans $\mathcal H$, the matrix $(L_{\mathcal H})_{ij}=\sqrt{q_iq_j}\,b_i^\top b_j$ combines plan quality with similarity between prediction changes. We reward the increase in quality-weighted coverage when new plans are added to the archive
\begin{small}
\begin{equation}
  D=\frac{\log\det(I+L_{\mathcal H_s\cup\mathcal P})
              -\log\det(I+L_{\mathcal H_s})}{J\log2},
  \qquad U=0.75Q+0.25D.
  \label{eq:useful-diversity}
\end{equation}
\end{small}
Here $I$ is the identity matrix. High-quality plans receive more weight, while similar prediction changes receive diminishing reward. Subtracting the archive term measures the gain beyond earlier research, and $J\log2$ normalizes $D$ to $[0,1]$. The shared utility $U$ therefore favors predictive quality and complementary, useful contributions. Please see Appendix~\ref{app:diversity-reward-details} for examples illustrating the effect of repeated directions and historical coverage.

\textbf{Role-specific credit and policy updates.}
The Planner chooses research directions, while the Realizer implements a given set of plans. We reflect this distinction when assigning credit from the shared utility. Let $U_{mn}$ be the utility of $(\mathcal P_m,\mathcal Z_{mn})$ and $\bar U_m=N^{-1}\sum_nU_{mn}$. The Planner compares a portfolio's average utility with the averages of other portfolios. The Realizer compares an implementation with alternatives generated for the same plans and retrieved content
\begin{small}
\begin{equation}
  A_m^{\mathrm P}=
    \frac{\bar U_m-\frac{1}{M-1}\sum_{m'\ne m}\bar U_{m'}}{c_{\mathrm P}},
  \qquad
  A_{mn}^{\mathrm R}=
    \frac{U_{mn}-\frac{1}{N-1}\sum_{n'\ne n}U_{mn'}}{c_{\mathrm R}}.
  \label{eq:conditional-advantages}
\end{equation}
\end{small}
The fixed positive scales $c_{\mathrm P}$ and $c_{\mathrm R}$ calibrate the two roles' advantages. For role $a$, let $A^a$ be the response advantage and $\rho_\ell^a$ the token-probability ratio between the current policy and the policy used for sampling. We update each policy using the clipped GRPO objective
\begin{small}
\begin{equation}
  \mathcal L_{\mathrm{RL}}(\theta_a)=
  -\mathbb E\!\left[\frac{1}{|v|}\sum_{\ell=1}^{|v|}
    \min\!\left(\rho_\ell^a A^a,
      \operatorname{clip}(\rho_\ell^a,1-\epsilon,1+\epsilon)A^a\right)\right]
    +\beta_a\mathcal L_{\mathrm{KL}}^a,
  \label{eq:joint-grpo}
\end{equation}
\end{small}
where $\epsilon$ controls clipping and $\mathcal L_{\mathrm{KL}}^a$ penalizes divergence from the frozen SFT policy with weight $\beta_a$. Positive advantages increase the probability of better plans or implementations under their respective comparisons. Only the two LoRA adapters are updated. Please see Appendix~\ref{app:invalid-rollouts} for detailed optimization settings.

\begingroup
\setlength{\textfloatsep}{12pt plus 2pt minus 2pt}
\setlength{\floatsep}{10pt plus 2pt minus 2pt}
\setlength{\intextsep}{10pt plus 2pt minus 2pt}
\section{Experiments}
\label{sec:experiments}

\textbf{Data and evaluation.}
We study alpha factor mining on CSI300, CSI500, CSI1000, and the broader Shanghai/Shenzhen A-share universe using daily and minute-level price and volume data. The prediction target is excess return from open $t+1$ to open $t+2$. The inner period, May 2022--December 2025, supports research and model selection; January--August 2026 is used once for final outer evaluation. Agentic methods select factors on inner feedback and refit the predictive model before outer evaluation. Please see Appendix~\ref{app:data-and-evaluation} for the data and evaluation details.

\textbf{Metrics and baselines.}
We report IC, ICIR, RIC, and RICIR for predictive quality, and ARR, IR, MDD, and CR for portfolio performance. \textbf{ML baselines} are Ridge and MLP~\citep{gu2020empirical}, and LightGBM~\citep{ke2017lightgbm}. \textbf{DL baselines} include GRU~\citep{cho2014gru}, LSTM~\citep{hochreiter1997lstm}, ALSTM~\citep{alstm}, TCN~\citep{bai2018tcn}, Transformer~\citep{vaswani2017attention}, PatchTST~\citep{nie2023patchtst}, iTransformer~\citep{liu2024itransformer}, MASTER~\citep{li2024master}, and StockMixer~\citep{fan2024stockmixer}. \textbf{Agentic comparisons} use RD-Agent(Q)~\citep{li2025rdagent}, AlphaAgent~\citep{tang2025alphaagent}, QuantaAlpha~\citep{han2026quantaalpha}, and AlphaSchema~\citep{alphaschema}. Please see Appendix~\ref{app:metric-definitions} for detailed definitions of the metrics, and Appendix~\ref{app:baseline-reproduction} for detailed descriptions of the baselines and their settings.

\textbf{Research models.}
AlphaDiverse uses post-trained Qwen3.8-27B Planner and Realizer policies and an unmodified Qwen3.8-27B Analysis model. Other agentic baselines use GPT-5.5. We also evaluate API-only AlphaDiverse with GPT-5.5, Grok-4.6, and GLM-5.3. All evaluation inference uses temperature 0.1. Each run starts from $|\mathcal B|=21$ factors and uses $R=20$ rounds, $J=4$ plans per round, and two specifications per plan, totaling 160 candidate slots. Main comparisons use all available features and $K=1$ inner fold, with $m_K=1$. Please see Appendix~\ref{app:models-hardware} for the training and serving settings.

\begin{table}[!t]
\centering
\caption{Experiment results about CSI300 predictive and portfolio performance. ARR/MDD are percentages; best and second-best values are bold and underlined (lower MDD is better). Other markets and API backends appear in Appendices~\ref{app:additional-results} and~\ref{app:api-results}.}
\label{tab:main-results}
\footnotesize\setlength{\tabcolsep}{2.4pt}\renewcommand{\arraystretch}{1.06}
\begin{tabular}{llrrrrrrrr}
\toprule
\multirow{2}{*}{Category} & \multirow{2}{*}{Method} & \multicolumn{4}{c}{Predictive performance} & \multicolumn{4}{c}{Portfolio performance} \\
\cmidrule(lr){3-6}\cmidrule(lr){7-10}
& & IC$\uparrow$ & ICIR$\uparrow$ & RIC$\uparrow$ & RICIR$\uparrow$ & ARR$\uparrow$ & IR$\uparrow$ & MDD$\downarrow$ & CR$\uparrow$ \\
\midrule
 & Ridge & 0.0166 & 0.0865 & 0.0210 & 0.1184 & 14.44 & 0.889 & 13.49 & 1.071 \\
 & MLP & 0.0287 & 0.1320 & 0.0377 & 0.1905 & 27.34 & 1.604 & 13.34 & 2.050 \\
\multirow{-3}{*}{ML} & LightGBM & \underline{0.0348} & 0.2043 & 0.0335 & 0.2243 & 24.15 & 1.846 & 6.99 & 3.456 \\
\midrule
 & GRU & 0.0303 & 0.1165 & 0.0170 & 0.0720 & 18.46 & 1.024 & 13.81 & 1.316 \\
 & LSTM & \underline{0.0348} & 0.1460 & 0.0277 & 0.1192 & 24.83 & 1.280 & 16.50 & 1.505 \\
 & ALSTM & 0.0181 & 0.0964 & 0.0296 & 0.1539 & 25.23 & 1.540 & 8.93 & 2.826 \\
 & TCN & 0.0131 & 0.0664 & 0.0212 & 0.1229 & 16.70 & 1.140 & 9.32 & 1.793 \\
 & Transformer & 0.0134 & 0.0607 & 0.0353 & 0.1492 & 7.45 & 0.439 & 13.78 & 0.500 \\
 & PatchTST & 0.0145 & 0.0600 & -0.0068 & -0.0259 & -3.59 & -0.274 & 21.19 & -0.310 \\
 & iTransformer & 0.0212 & 0.1473 & 0.0171 & 0.1284 & 15.61 & 0.857 & 6.76 & 1.471 \\
 & MASTER & 0.0318 & 0.1516 & 0.0325 & 0.1974 & 27.02 & 1.858 & 10.63 & 2.265 \\
\multirow{-9}{*}{DL} & StockMixer & -0.0069 & -0.0320 & 0.0236 & 0.1019 & -4.75 & -0.240 & 22.72 & -0.226 \\
\midrule
 & RD-Agent(Q) & 0.0244 & 0.1299 & 0.0223 & 0.1342 & 17.57 & 1.355 & 10.03 & 1.751 \\
 & AlphaAgent & 0.0337 & \underline{0.2075} & 0.0384 & \underline{0.2587} & 33.62 & 2.856 & \underline{6.05} & 5.554 \\
 & QuantaAlpha & 0.0340 & 0.1804 & \underline{0.0403} & 0.2449 & \textbf{43.19} & \underline{2.980} & 6.51 & \underline{6.634} \\
 & AlphaSchema & 0.0338 & 0.1965 & 0.0367 & 0.2156 & 32.90 & 2.272 & 7.61 & 4.324 \\
\rowcolor{MethodTint}
\multirow{-5}{*}{Agentic} & AlphaDiverse & \textbf{0.0378} & \textbf{0.2128} & \textbf{0.0407} & \textbf{0.2790} & \underline{42.10} & \textbf{3.210} & \textbf{5.51} & \textbf{7.641} \\
\bottomrule
\end{tabular}
\end{table}

\begin{table}[!t]\centering
\begin{minipage}[t]{0.65\linewidth}\vspace{0pt}
\captionsetup{font=small,skip=3pt}
\caption{Results of diversity on CSI300 and CSI500 across agentic methods.}\label{tab:research-diversity}
\footnotesize\setlength{\tabcolsep}{1.7pt}\renewcommand{\arraystretch}{1.03}
\begin{tabular*}{\linewidth}{@{\extracolsep{\fill}}lrrrrrr}\toprule
\multirow{2}{*}{Method} & \multicolumn{3}{c}{CSI300} & \multicolumn{3}{c}{CSI500} \\
\cmidrule(lr){2-4}\cmidrule(lr){5-7}
 & M$\uparrow$ & U$\uparrow$ & C$\uparrow$ & M$\uparrow$ & U$\uparrow$ & C$\uparrow$ \\ \midrule
RD-Agent(Q) & 18 & 3 & 67 & \underline{25} & \underline{14} & 121 \\
AlphaAgent & 5 & 5 & 16 & 5 & 1 & 34 \\
QuantaAlpha & 11 & 3 & 58 & 22 & 4 & 64 \\
AlphaSchema & 18 & 2 & 96 & 19 & 7 & 96 \\
Single Synthesis & 19 & 8 & \underline{122} & 20 & 10 & 99 \\
AlphaDiverse (GPT-5.5) & 24 & \underline{11} & 108 & 24 & 11 & \underline{125} \\
AlphaDiverse (SFT-only) & \underline{27} & 4 & 85 & 23 & 6 & 83 \\
\rowcolor{MethodTint}
AlphaDiverse & \textbf{34} & \textbf{23} & \textbf{129} & \textbf{36} & \textbf{25} & \textbf{128} \\
\bottomrule\end{tabular*}
\end{minipage}\hfill
\begin{minipage}[t]{0.33\linewidth}\vspace{0pt}\centering
\includegraphics[width=\linewidth]{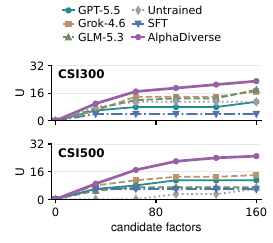}
\captionsetup{font=small,skip=3pt}
\captionof{figure}{Useful exploration among different AlphaDiverse variants.}\label{fig:research-diversity}
\end{minipage}\end{table}

\subsection{Main Results}
\label{sec:main-performance}
To evaluate the financial performance of post-trained local research policies, we compare AlphaDiverse with ML, DL, and agentic baselines under the same evaluation protocol. As Table~\ref{tab:main-results} shows, AlphaDiverse leads all four predictive metrics, including IC of 0.0378 and RIC of 0.0407. It also achieves the highest IR (3.210) and CR (7.641), the lowest MDD (5.51\%), and the second-highest ARR (42.10\%). These results support strong predictive and risk-adjusted portfolio performance with local research policies. Results of all four markets in Appendix~\ref{app:additional-results} exhibit the same pattern.

\subsection{Research Diversity and Generalization}
\label{sec:diversity-experiments}

We propose four metrics to evaluate the diversity of research. (1) \emph{Mechanisms} (M) counts catalogue entries represented by generated factors; (2) \emph{useful mechanisms} (U) counts those with model-retained factors. (3) \emph{Signal clusters} (C) counts numerically distinct signal groups. (4) \emph{Pair corr} is the median absolute correlation between the two factors realizing one plan for AlphaDiverse variants with paired implementations only; lower values indicate more distinct implementations.
Please see Appendix~\ref{app:diversity-protocol} for detailed definitions of the diversity metrics.

\begin{table}[!t]\centering
\begin{minipage}[t]{0.49\linewidth}\vspace{0pt}\centering
\caption{Results of Module ablations (GPT-5.5).}\label{tab:ablation}
\footnotesize\setlength{\tabcolsep}{1.2pt}\renewcommand{\arraystretch}{1.08}
\begin{tabular}{lrrrrr}\toprule
Variant & RIC$\uparrow$ & ARR$\uparrow$ & M$\uparrow$ & U$\uparrow$ & Pair$\downarrow$ \\ 
\midrule
- w/o complementarity & 0.0352 & \textbf{37.73} & 10 & 7 & 0.455 \\
- w/o retrieval & 0.0373 & 28.39 & \textbf{25} & \underline{9} & \underline{0.371} \\
- w/o memory & \underline{0.0383} & 32.26 & 21 & \textbf{11} & 0.473 \\
Single Synthesis & 0.0316 & 23.83 & 19 & 8 & --- \\
\rowcolor{MethodTint}
AlphaDiverse (GPT-5.5) & \textbf{0.0398} & \underline{35.97} & \underline{24} & \textbf{11} & \textbf{0.322} \\
\bottomrule\end{tabular}

\end{minipage}\hfill
\begin{minipage}[t]{0.49\linewidth}\vspace{0pt}\centering

\caption{Results of Post-training ablations.}\label{tab:training-ablation}
\footnotesize\setlength{\tabcolsep}{1.2pt}\renewcommand{\arraystretch}{1.08}
\begin{tabular}{lrrrrr}\toprule
Variant & RIC$\uparrow$ & ARR$\uparrow$ & M$\uparrow$ & U$\uparrow$ & Pair$\downarrow$ \\
\midrule
- w/o training & 0.0291 & 22.49 & \underline{33} & 11 & 0.413 \\
- w/ SFT & 0.0389 & 24.77 & 27 & 4 & 0.511 \\
- w/ Planner only & 0.0397 & 32.57 & 30 & 13 & 0.423 \\
- w/ Realizer only & 0.0400 & 35.17 & 30 & 15 & 0.394 \\
- w/o diversity reward & \underline{0.0403} & \underline{37.77} & 29 & \underline{18} & \underline{0.365} \\
\rowcolor{MethodTint}
AlphaDiverse & \textbf{0.0407} & \textbf{42.10} & \textbf{34} & \textbf{23} & \textbf{0.316} \\
\bottomrule\end{tabular}
\end{minipage}
\end{table}

\textit{Verifying whether AlphaDiverse produces diverse research paths.} We compare agentic baselines with complete AlphaDiverse and multiple AlphaDiverse variants, including Single Synthesis (replaces the Planner--Realizer decomposition in Section~\ref{sec:agent-system} with one agent that directly proposes eight factors), API-based (GPT-5.5), and SFT-only AlphaDiverse.
As Table~\ref{tab:research-diversity} shows, complete AlphaDiverse consistently outperforms all baselines in three diversity metrics, and increases useful mechanisms from 4 to 23 and from 6 to 25 relative to SFT. The comparison with Single Synthesis is instructive: on CSI300 it generates nearly as many clusters, yet far fewer retained mechanisms. Organizing research around explicit plans helps turn varied implementations into economically distinct model inputs. Results of all four markets in Appendix~\ref{app:cross-method-diversity} exhibit the same pattern.

\textit{Verifying whether useful mechanisms are proposed continuously.} We compare multiple AlphaDiverse variants, including API-based (GPT-5.5, Grok-4.6 and GLM-5.3), untrained (using the original Qwen3.8-27B), and SFT-only AlphaDiverse. As Figure~\ref{fig:research-diversity} shows, we continuously monitor the number of useful mechanisms discovered by the methods within a loop. SFT saturates early in both markets, whereas AlphaDiverse continues to add mechanisms as the budget grows. This supports sustained discovery under feedback: later proposals still alter the retained model. Results of all four markets in Appendix~\ref{app:coverage-results} exhibit the same pattern.

\begin{figure}[!ht]\centering
\begin{minipage}[t]{0.63\linewidth}\vspace{0pt}
\textit{Verifying whether diverse research capabilities generalize.} We compare API-based (GPT-5.5), SFT-only, and complete AlphaDiverse on shared states from historical runs of all three APIs. For each market, we sample 10 states from runs used to construct training data (Familiar) and 10 from excluded runs (Held-out). Each variant performs eight independent single-round experiments from every fixed state. Figure~\ref{fig:fixed-state} shows that AlphaDiverse retains more useful mechanisms in both groups. This supports AlphaDiverse's ability to explore useful alternatives in unseen research histories. Appendix~\ref{app:coverage-results} provides detailed settings and complete results, which exhibit the same pattern.
\end{minipage}\hfill
\begin{minipage}[t]{0.34\linewidth}\vspace{0pt}\centering
\includegraphics[width=\linewidth]{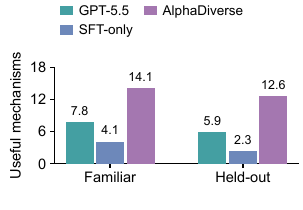}
\captionsetup{font=small,skip=3pt}
\caption{Results of Fixed-state transfer test.}\label{fig:fixed-state}
\end{minipage}\end{figure}
\vspace{-10pt}

\subsection{Ablation Study}
\label{sec:experiment-ablations}
\textbf{Module ablations.} To test the effect of each module in the proposed agent system, we evaluate four ablation variants for modules in Section~\ref{sec:agent-system}: (1) \emph{w/o complementarity} (remove Planner complementarity rules); (2) \emph{w/o retrieval} (replace plan-specific retrieval with all legal features and operators); (3) \emph{w/o memory} (do not provide the Planner with full research history but only the retained factors and performance); and (4) \emph{Single Synthesis} (replace the Planner and Realizer with Single Synthesis). Execution, validation, Analysis, and the GPT-5.5 backend remain fixed. As Table~\ref{tab:ablation} shows, removing complementarity permits a slightly higher ARR but concentrates research on only 10 mechanisms. Removing retrieval preserves breadth yet lowers useful coverage and predictive quality, meaning that an idea still needs relevant inputs to become a useful factor. Removing memory weakens prediction performance. Together, these results support a division of labor: planning broadens directions, retrieval supports implementation, and feedback improves choices within those directions. Appendix~\ref{app:ablation-interventions} defines the detailed ablation setting, and results for all four markets in Table~\ref{tab:workflow-complete} exhibit the same pattern. Appendix~\ref{app:factor-quality} reports factor quality under different AlphaDiverse settings.

\begin{table}[!t]\centering
\begin{minipage}[t]{0.49\linewidth}\vspace{0pt}\centering

\caption{SFT-data ablations.}\label{tab:data-ablation}
\footnotesize\setlength{\tabcolsep}{1pt}\renewcommand{\arraystretch}{1.08}
\begin{tabular}{lrrrrr}\toprule
Data selection & RIC$\uparrow$ & ARR$\uparrow$ & M$\uparrow$ & U$\uparrow$ & Pair$\downarrow$ \\
\midrule
- w/ improvement only & 0.0371 & 22.38 & 21 & 1 & 0.616 \\
- w/o balancing & \underline{0.0383} & \underline{23.95} & \underline{25} & \underline{3} & \underline{0.546} \\
- w/o paired supervision & 0.0365 & 21.51 & 19 & 1 & 0.651 \\
\rowcolor{MethodTint}
full data SFT & \textbf{0.0389} & \textbf{24.77} & \textbf{27} & \textbf{4} & \textbf{0.511} \\
\bottomrule\end{tabular}

\end{minipage}\hfill
\begin{minipage}[t]{0.49\linewidth}\vspace{0pt}\centering
\caption{Deployment costs on CSI300.}\label{tab:deployment-cost}
\footnotesize\setlength{\tabcolsep}{2pt}\renewcommand{\arraystretch}{1.12}
{
\begin{tabular*}{\linewidth}{@{\extracolsep{\fill}}lrr}\toprule
Deployment & API fees (\$) & GPU-hours \\ \midrule
GPT-5.5 & 7.28 & 0.000 \\
Grok-4.6 & 5.57 & 0.000 \\
GLM-5.3 & 2.19 & 0.000 \\
\rowcolor{MethodTint}
AlphaDiverse (local) & 0.00 & 1.436 \\
\bottomrule\end{tabular*}}
\end{minipage}
\end{table}

\textbf{Post-training ablations.} To further evaluate our post-training pipeline in Section~\ref{sec:local-post-training}, we compare five post-training settings: (1) \emph{w/o training} (untrained original model); (2) \emph{w/ SFT}; (3) \emph{w/ Planner only } (frozen SFT Realizer in joint GRPO); (4) \emph{w/ Realizer only } (frozen SFT Planner in joint GRPO); and (5) \emph{w/o diversity reward} (joint GRPO without $D$ in Eq.~\ref{eq:useful-diversity}). As Table~\ref{tab:training-ablation} shows, SFT improves prediction but narrows useful coverage. Single-role training recovers part of the gain. Joint training aligns plans with implementations. The diversity term further increases coverage (U: 18 to 23) and reduces pair correlation, identifying a benefit beyond joint quality optimization. Appendix~\ref{app:ablation-interventions} defines the detailed ablation setting, and results for all four markets in Table~\ref{tab:objective-complete} exhibit the same conclusion.

\textbf{Data ablations.} To evaluate our data selection strategy in Section~\ref{sec:training-data}, we compare four data selection settings and evaluate them at the SFT stage: (1) \emph{w/ improvement only}; (2) \emph{w/o balancing} (no within-group balancing across environments); (3) \emph{w/o paired supervision} (no pair-aware Realizer supervision); and (4) \emph{full data SFT}. As Table~\ref{tab:data-ablation} shows, improvement-only selection loses useful mechanisms, indicating that successful endpoints omit informative changes of direction. Removing paired supervision increases implementation correlation (0.511 to 0.651) and reduces retained coverage. The Planner needs varied decisions; the Realizer needs distinct implementations. Appendix~\ref{app:ablation-interventions} defines the ablation setting, and 
results for all four markets in Table~\ref{tab:data-complete} exhibit the same pattern.

\textbf{Deployment cost.} To evaluate the deployment cost of AlphaDiverse, we compare one complete 20-round CSI300 run across three API backends and our local deployment. As Table 6 shows, the API-based runs incur estimated charges of \$7.28, \$5.57, and \$2.19 for GPT-5.5, Grok-4.6, and GLM-5.3, respectively. Local AlphaDiverse incurs no external API charges and uses 1.436 H200 GPU-hours. It also supports flexible deployment on compatible, lower-cost GPUs. Combined with its stronger overall performance than all three API variants across four markets, these results demonstrate competitive local research without API fees or dependence on external API availability. Token details appear in Appendix~\ref{app:deployment-details}.

\section{Related Work}
\label{sec:related-work}

\noindent\textbf{LLM-Driven Financial Agents.}
LLMs support financial prediction and trading through training for financial tasks \citep{yang2023fingpt,xiong2025flagtrader,deng2026alphaquanter}. Multi-agent systems assign research planning, implementation, and evaluation to distinct agents \citep{xiao2024tradingagents,li2025rdagent,guo2026aqua}. Recent work also trains alpha generation policies or uses external memory to guide search \citep{tang2026alphaagentevo,zhang2026quantevolver,factorminer,alphamemo}. 

\noindent\textbf{Research-Path Diversity.}
Existing methods encourage exploration through expression regularization, evolutionary search, and control of factor redundancy \citep{tang2025alphaagent,liu2026cognitive,alphamaster}. Semantic planning makes the economic ideas behind factors explicit \citep{alphaschema}, while studies of idea generation show that distinct outputs can still converge on similar research directions \citep{audranreiss2025ideation,chen2026diversitycollapse}. We require different economic mechanisms in each round and collect traces across research environments to train agents for diverse exploration. Appendix~\ref{app:additional-related-work} provides further discussion of broader related work.

\section{Conclusion and Limitations}
\label{sec:conclusion}
In this paper, we propose AlphaDiverse, a framework for training local quantitative research agents for diverse alpha factor mining. Complementary planning across varied research environments produces diverse traces for supervised fine-tuning. Joint reinforcement learning then optimizes the Planner and Realizer for predictive quality and useful diversity. With research feedback separated from outer evaluation, experiments across four stock universes show competitive financial performance and sustained exploration, supporting effective alpha research with locally deployed agents.

\textbf{Limitations.}
The limitations of the proposed AlphaDiverse are as follows:  (1) the fixed mechanism catalogue and operator library bound the ideas that agents can express; extending them requires reliable semantic and causal checks; (2) joint training evaluates sampled plans on a finite bank of inner states, which only approximates their delayed contribution during a long research loop. Learning to expand the search language and assigning credit across evolving research states are promising directions for further work.

\label{end:main-text}
\endgroup

\bibliography{references}

\begin{thebibliography}{56}
\providecommand{\natexlab}[1]{#1}
\providecommand{\url}[1]{\texttt{#1}}
\expandafter\ifx\csname urlstyle\endcsname\relax
  \providecommand{\doi}[1]{doi: #1}\else
  \providecommand{\doi}{doi: \begingroup \urlstyle{rm}\Url}\fi

\bibitem[Audran-Reiss et~al.(2025)Audran-Reiss, Armengol~Estap{\'{e}},
  Hambardzumyan, Budhiraja, Josifoski, Toledo, Hazra, Magka, Shvartsman,
  Pathak, Kao, Cipolina-Kun, Gauri, Gagnon-Audet, Tewolde, Zhang, Cohen, Adi,
  Shavrina, and Bachrach]{audranreiss2025ideation}
Alexis Audran-Reiss, Jordi Armengol~Estap{\'{e}}, Karen Hambardzumyan, Amar
  Budhiraja, Martin Josifoski, Edan Toledo, Rishi Hazra, Despoina Magka,
  Michael Shvartsman, Parth Pathak, Justine~T. Kao, Lucia Cipolina-Kun, Bhavul
  Gauri, Jean-Christophe Gagnon-Audet, Emanuel Tewolde, Jenny Zhang, Taco
  Cohen, Yossi Adi, Tatiana Shavrina, and Yoram Bachrach.
\newblock What does it take to be a good {AI} research agent? studying the role
  of ideation diversity.
\newblock \emph{arXiv preprint arXiv:2511.15593}, 2025.
\newblock URL \url{https://arxiv.org/abs/2511.15593}.

\bibitem[Bai et~al.(2018)Bai, Kolter, and Koltun]{bai2018tcn}
Shaojie Bai, J.~Zico Kolter, and Vladlen Koltun.
\newblock An empirical evaluation of generic convolutional and recurrent
  networks for sequence modeling.
\newblock \emph{arXiv preprint arXiv:1803.01271}, 2018.
\newblock URL \url{https://arxiv.org/abs/1803.01271}.

\bibitem[Cawley \& Talbot(2010)Cawley and Talbot]{cawley2010overfitting}
Gavin~C. Cawley and Nicola L.~C. Talbot.
\newblock On over-fitting in model selection and subsequent selection bias in
  performance evaluation.
\newblock \emph{Journal of Machine Learning Research}, 11:\penalty0 2079--2107,
  2010.
\newblock URL \url{https://www.jmlr.org/papers/v11/cawley10a.html}.

\bibitem[Chen et~al.(2024)Chen, Pelger, and Zhu]{chen2024deeplearning}
Luyang Chen, Markus Pelger, and Jason Zhu.
\newblock Deep learning in asset pricing.
\newblock \emph{Management Science}, 70\penalty0 (2):\penalty0 714--750, 2024.
\newblock \doi{10.1287/mnsc.2023.4695}.

\bibitem[Chen et~al.(2026)Chen, Tong, Yang, He, Zhang, Zou, Wang, and
  He]{chen2026diversitycollapse}
Nuo Chen, Yicheng Tong, Yuzhe Yang, Yufei He, Xueyi Zhang, Qingyun Zou, Qian
  Wang, and Bingsheng He.
\newblock Diversity collapse in multi-agent {LLM} systems: Structural coupling
  and collective failure in open-ended idea generation.
\newblock In \emph{Findings of the Association for Computational Linguistics:
  ACL 2026}, 2026.
\newblock URL \url{https://arxiv.org/abs/2604.18005}.

\bibitem[Cho et~al.(2014)Cho, van Merrienboer, Gulcehre, Bahdanau, Bougares,
  Schwenk, and Bengio]{cho2014gru}
Kyunghyun Cho, Bart van Merrienboer, Caglar Gulcehre, Dzmitry Bahdanau, Fethi
  Bougares, Holger Schwenk, and Yoshua Bengio.
\newblock Learning phrase representations using {RNN} encoder--decoder for
  statistical machine translation.
\newblock In \emph{Proceedings of the Conference on Empirical Methods in
  Natural Language Processing}, 2014.
\newblock URL \url{https://arxiv.org/abs/1406.1078}.

\bibitem[Deng et~al.(2026)Deng, Yan, Yu, and Wang]{deng2026alphaquanter}
Zheye Deng, Weixiang Yan, Changlong Yu, and Jiashu Wang.
\newblock {AlphaQuanter}: An end-to-end tool-augmented agentic reinforcement
  learning framework for stock trading.
\newblock In \emph{Findings of the Association for Computational Linguistics:
  ACL 2026}, pp.\  9373--9394, 2026.
\newblock \doi{10.18653/v1/2026.findings-acl.456}.
\newblock URL \url{https://aclanthology.org/2026.findings-acl.456/}.

\bibitem[Duan et~al.(2022)Duan, Wang, Zhang, and Li]{duan2022factorvae}
Yitong Duan, Lei Wang, Qizhong Zhang, and Jian Li.
\newblock {FactorVAE}: A probabilistic dynamic factor model based on
  variational autoencoder for predicting cross-sectional stock returns.
\newblock In \emph{Proceedings of the AAAI Conference on Artificial
  Intelligence}, volume~36, pp.\  4468--4476, 2022.
\newblock \doi{10.1609/aaai.v36i4.20369}.

\bibitem[Duan et~al.(2025)Duan, Wang, and Li]{duan2025factorgcl}
Yitong Duan, Weiran Wang, and Jian Li.
\newblock {FactorGCL}: A hypergraph-based factor model with temporal residual
  contrastive learning for stock returns prediction.
\newblock In \emph{Proceedings of the AAAI Conference on Artificial
  Intelligence}, volume~39, pp.\  173--181, 2025.
\newblock \doi{10.1609/aaai.v39i1.31993}.

\bibitem[Dwork et~al.(2015)Dwork, Feldman, Hardt, Pitassi, Reingold, and
  Roth]{dwork2015adaptive}
Cynthia Dwork, Vitaly Feldman, Moritz Hardt, Toni Pitassi, Omer Reingold, and
  Aaron Roth.
\newblock Generalization in adaptive data analysis and holdout reuse.
\newblock In \emph{Advances in Neural Information Processing Systems},
  volume~28, 2015.
\newblock URL
  \url{https://proceedings.neurips.cc/paper/2015/hash/bad5f33780c42f2588878a9d07405083-Abstract.html}.

\bibitem[Fan \& Shen(2024)Fan and Shen]{fan2024stockmixer}
Jinyong Fan and Yanyan Shen.
\newblock {StockMixer}: A simple yet strong {MLP}-based architecture for stock
  price forecasting.
\newblock In \emph{Proceedings of the AAAI Conference on Artificial
  Intelligence}, volume~38, pp.\  8389--8397, 2024.
\newblock \doi{10.1609/aaai.v38i8.28681}.
\newblock URL \url{https://ojs.aaai.org/index.php/AAAI/article/view/28681}.

\bibitem[Foerster et~al.(2018)Foerster, Farquhar, Afouras, Nardelli, and
  Whiteson]{foerster2018coma}
Jakob Foerster, Gregory Farquhar, Triantafyllos Afouras, Nantas Nardelli, and
  Shimon Whiteson.
\newblock Counterfactual multi-agent policy gradients.
\newblock In \emph{Proceedings of the AAAI Conference on Artificial
  Intelligence}, volume~32, 2018.
\newblock \doi{10.1609/aaai.v32i1.11794}.
\newblock URL \url{https://ojs.aaai.org/index.php/AAAI/article/view/11794}.

\bibitem[Gu et~al.(2020)Gu, Kelly, and Xiu]{gu2020empirical}
Shihao Gu, Bryan Kelly, and Dacheng Xiu.
\newblock Empirical asset pricing via machine learning.
\newblock \emph{The review of financial studies}, 33\penalty0 (5):\penalty0
  2223--2273, 2020.

\bibitem[Guo et~al.(2026)Guo, Huang, Gao, Li, Ge, Kuang, and Wang]{guo2026aqua}
Jiacheng Guo, Suozhi Huang, Yunlong Gao, Zihao Li, Jason Ge, Xu~Kuang, and
  Mengdi Wang.
\newblock {AQuA}: Recursively self-improving quantitative trading research
  agents.
\newblock \emph{arXiv preprint arXiv:2608.12841}, 2026.
\newblock URL \url{https://arxiv.org/abs/2608.12841}.

\bibitem[Han et~al.(2026)Han, Zhang, Li, Dong, Hu, Zhu, Yu, Guo, Liu, Wang,
  et~al.]{han2026quantaalpha}
Jun Han, Shuo Zhang, Wei Li, Yifan Dong, Tu~Hu, Yumo Zhu, Xiaomin Yu, Xin Guo,
  Zhaowei Liu, Kunyi Wang, et~al.
\newblock Quantaalpha: An evolutionary framework for llm-driven alpha mining.
\newblock \emph{arXiv preprint arXiv:2602.07085}, 2026.
\newblock URL \url{https://arxiv.org/abs/2602.07085}.

\bibitem[Hochreiter \& Schmidhuber(1997)Hochreiter and
  Schmidhuber]{hochreiter1997lstm}
Sepp Hochreiter and J{\"u}rgen Schmidhuber.
\newblock Long short-term memory.
\newblock \emph{Neural Computation}, 9\penalty0 (8):\penalty0 1735--1780, 1997.
\newblock \doi{10.1162/neco.1997.9.8.1735}.

\bibitem[Hong et~al.(2025)Hong, Yin, Wang, Liu, Chen, Yu, Li, Ye, Xiao, Chen,
  Zhou, Yue, Yang, Guo, Liu, Wei, and Gu]{hong2025mgrpo}
Haoyang Hong, Jiajun Yin, Yuan Wang, Jingnan Liu, Zhe Chen, Ailing Yu, Ji~Li,
  Zhiling Ye, Hansong Xiao, Yefei Chen, Hualei Zhou, Yun Yue, Minghui Yang,
  Chunxiao Guo, Junwei Liu, Peng Wei, and Jinjie Gu.
\newblock Multi-agent deep research: Training multi-agent systems with
  {M-GRPO}.
\newblock \emph{arXiv preprint arXiv:2511.13288}, 2025.
\newblock URL \url{https://arxiv.org/abs/2511.13288}.

\bibitem[Hu et~al.(2021)Hu, Shen, Wallis, Allen-Zhu, Li, Wang, Wang, and
  Chen]{hu2021lora}
Edward~J. Hu, Yelong Shen, Phillip Wallis, Zeyuan Allen-Zhu, Yuanzhi Li, Shean
  Wang, Lu~Wang, and Weizhu Chen.
\newblock {LoRA}: Low-rank adaptation of large language models.
\newblock \emph{arXiv preprint arXiv:2106.09685}, 2021.
\newblock URL \url{https://arxiv.org/abs/2106.09685}.

\bibitem[Ke et~al.(2017)Ke, Meng, Finley, Wang, Chen, Ma, Ye, and
  Liu]{ke2017lightgbm}
Guolin Ke, Qi~Meng, Thomas Finley, Taifeng Wang, Wei Chen, Weidong Ma, Qiwei
  Ye, and Tie-Yan Liu.
\newblock {LightGBM}: A highly efficient gradient boosting decision tree.
\newblock In \emph{Advances in Neural Information Processing Systems}, 2017.
\newblock URL
  \url{https://papers.nips.cc/paper/2017/hash/6449f44a102fde848669bdd9eb6b76fa-Abstract.html}.

\bibitem[Li et~al.(2024{\natexlab{a}})Li, Liu, Shen, Wang, Chen, and
  Huang]{li2024master}
Tong Li, Zhaoyang Liu, Yanyan Shen, Xue Wang, Haokun Chen, and Sen Huang.
\newblock {MASTER}: Market-guided stock transformer for stock price
  forecasting.
\newblock In \emph{Proceedings of the AAAI Conference on Artificial
  Intelligence}, volume~38, pp.\  162--170, 2024{\natexlab{a}}.
\newblock \doi{10.1609/aaai.v38i1.27767}.

\bibitem[Li et~al.(2026{\natexlab{a}})Li, Yang, Yang, Wang, Liu, and
  Bian]{li2025rdagent}
Yuante Li, Xu~Yang, Xiao Yang, Xisen Wang, Weiqing Liu, and Jiang Bian.
\newblock R\&d-agent-quant: A multi-agent framework for data-centric factors
  and model joint optimization.
\newblock In \emph{The Thirty-ninth Annual Conference on Neural Information
  Processing Systems Datasets and Benchmarks Track}, 2026{\natexlab{a}}.
\newblock URL \url{https://openreview.net/forum?id=9VxTXAUH7G}.

\bibitem[Li et~al.(2024{\natexlab{b}})Li, Song, Sun, Xu, Yu, and
  Wen]{fama_agent}
Zhiwei Li, Ran Song, Caihong Sun, Wei Xu, Zhengtao Yu, and Ji-Rong Wen.
\newblock Can large language models mine interpretable financial factors more
  effectively? a neural-symbolic factor mining agent model.
\newblock In \emph{Findings of the Association for Computational Linguistics:
  ACL 2024}, pp.\  3891--3902, 2024{\natexlab{b}}.
\newblock URL \url{https://aclanthology.org/2024.findings-acl.233/}.

\bibitem[Li et~al.(2026{\natexlab{b}})Li, Li, Guo, Zhang, Collier, and
  Ie]{li2026autoscientistquant}
Zongqian Li, Yaoyiran Li, Yaohui Guo, Ming Zhang, Nigel Collier, and Eugene Ie.
\newblock {AutoScientist-Quant}: Self-evolving coding agents for automatic
  research in quantitative investment.
\newblock \emph{arXiv preprint arXiv:2608.28632}, 2026{\natexlab{b}}.
\newblock \doi{10.48550/arXiv.2608.28632}.
\newblock URL \url{https://arxiv.org/abs/2608.28632}.

\bibitem[Liao et~al.(2025)Liao, Wen, Wang, and Zhang]{liao2025marft}
Junwei Liao, Muning Wen, Jun Wang, and Weinan Zhang.
\newblock {MARFT}: Multi-agent reinforcement fine-tuning.
\newblock \emph{arXiv preprint arXiv:2504.16129}, 2025.
\newblock URL \url{https://arxiv.org/abs/2504.16129}.

\bibitem[Liu et~al.(2026{\natexlab{a}})Liu, Fu, Wang, and Liu]{xalpha}
Fengyuan Liu, Yuchen Fu, Yuqi Wang, and Qi~Liu.
\newblock Xalpha: A memory-driven ai quant researcher for hypothesis-to-code
  alpha discovery.
\newblock \emph{arXiv preprint arXiv:2607.08332}, 2026{\natexlab{a}}.
\newblock URL \url{https://arxiv.org/abs/2607.08332}.

\bibitem[Liu et~al.(2026{\natexlab{b}})Liu, Huang, Luo, Wang, Yang, Li, Hu,
  Feng, and Liu]{liu2026cognitive}
Fengyuan Liu, Yi~Huang, Sichun Luo, Yuqi Wang, Yazheng Yang, Xinye Li, Zefa Hu,
  Junlan Feng, and Qi~Liu.
\newblock Cognitive alpha mining via llm-driven code-based evolution.
\newblock In \emph{Proceedings of the 64th Annual Meeting of the Association
  for Computational Linguistics (Volume 1: Long Papers)}, pp.\  11715--11749,
  2026{\natexlab{b}}.
\newblock URL \url{https://aclanthology.org/2026.acl-long.538/}.

\bibitem[Liu et~al.(2026{\natexlab{c}})Liu, Liang, Lyu, and
  Amato]{liu2025magrpo}
Shuo Liu, Zeyu Liang, Xueguang Lyu, and Christopher Amato.
\newblock Llm collaboration with multi-agent reinforcement learning.
\newblock In \emph{Proceedings of the AAAI Conference on Artificial
  Intelligence}, volume~40, pp.\  32150--32158, 2026{\natexlab{c}}.

\bibitem[Liu et~al.(2024)Liu, Hu, Zhang, Wu, Wang, Ma, and
  Long]{liu2024itransformer}
Yong Liu, Tengge Hu, Haoran Zhang, Haixu Wu, Shiyu Wang, Lintao Ma, and
  Mingsheng Long.
\newblock {iTransformer}: Inverted transformers are effective for time series
  forecasting.
\newblock In \emph{International Conference on Learning Representations}, 2024.
\newblock URL \url{https://arxiv.org/abs/2310.06625}.

\bibitem[Lu et~al.(2024)Lu, Lu, Lange, Foerster, Clune, and
  Ha]{lu2024aiscientist}
Chris Lu, Cong Lu, Robert~Tjarko Lange, Jakob Foerster, Jeff Clune, and David
  Ha.
\newblock The {AI} scientist: Towards fully automated open-ended scientific
  discovery.
\newblock \emph{arXiv preprint arXiv:2408.06292}, 2024.
\newblock URL \url{https://arxiv.org/abs/2408.06292}.

\bibitem[Luo et~al.(2025)Luo, Kasirzadeh, and Shah]{luo2025pitfalls}
Ziming Luo, Atoosa Kasirzadeh, and Nihar~B. Shah.
\newblock The more you automate, the less you see: Hidden pitfalls of {AI}
  scientist systems.
\newblock \emph{arXiv preprint arXiv:2509.08713}, 2025.
\newblock URL \url{https://arxiv.org/abs/2509.08713}.

\bibitem[Nie et~al.(2023)Nie, Nguyen, Sinthong, and
  Kalagnanam]{nie2023patchtst}
Yuqi Nie, Nam~H. Nguyen, Phanwadee Sinthong, and Jayant Kalagnanam.
\newblock A time series is worth 64 words: Long-term forecasting with
  transformers.
\newblock In \emph{International Conference on Learning Representations}, 2023.
\newblock URL \url{https://arxiv.org/abs/2211.14730}.

\bibitem[Ning et~al.(2026{\natexlab{a}})Ning, Li, Zeng, Kang, and
  Xiong]{ning2026specialist}
Jingjie Ning, Xiaochuan Li, Ji~Zeng, Hao Kang, and Chenyan Xiong.
\newblock Auto research with specialist agents develops effective and
  non-trivial training recipes.
\newblock \emph{arXiv preprint arXiv:2605.05724}, 2026{\natexlab{a}}.
\newblock URL \url{https://arxiv.org/abs/2605.05724}.

\bibitem[Ning et~al.(2026{\natexlab{b}})Ning, Li, Zeng, Xiong, and
  Ke]{ning2026molprop}
Jingjie Ning, Xiaochuan Li, Ji~Zeng, Chenyan Xiong, and Guolin Ke.
\newblock Closed-loop auto research for molecular property prediction:
  Discovering and certifying generalizable improvements.
\newblock \emph{arXiv preprint arXiv:2606.22731}, 2026{\natexlab{b}}.
\newblock URL \url{https://arxiv.org/abs/2606.22731}.

\bibitem[Ning et~al.(2026{\natexlab{c}})Ning, Li, Zhong, Zeng, and
  Ke]{ning2026materials}
Jingjie Ning, Xiaochuan Li, Shanshan Zhong, Ji~Zeng, and Guolin Ke.
\newblock Auto research for materials: Auditable {AI}-scientist workflows with
  held-out transfer.
\newblock \emph{arXiv preprint arXiv:2607.17100}, 2026{\natexlab{c}}.
\newblock URL \url{https://arxiv.org/abs/2607.17100}.

\bibitem[Qian et~al.(2024)Qian, Zhou, Zhao, Chen, Yao, Wang, Liu, Yu, Zhang,
  and Zhou]{qian2024mdgnn}
Hao Qian, Hongting Zhou, Qian Zhao, Hao Chen, Hongxiang Yao, Jingwei Wang, Ziqi
  Liu, Fei Yu, Zhiqiang Zhang, and Jun Zhou.
\newblock {MDGNN}: Multi-relational dynamic graph neural network for
  comprehensive and dynamic stock investment prediction.
\newblock In \emph{Proceedings of the AAAI Conference on Artificial
  Intelligence}, volume~38, pp.\  14642--14650, 2024.
\newblock \doi{10.1609/aaai.v38i13.29381}.

\bibitem[Qin et~al.(2017)Qin, Song, Chen, Cheng, Jiang, and Cottrell]{alstm}
Yao Qin, Dongjin Song, Haifeng Chen, Wei Cheng, Guofei Jiang, and Garrison~W.
  Cottrell.
\newblock A dual-stage attention-based recurrent neural network for time series
  prediction.
\newblock In \emph{Proceedings of the Twenty-Sixth International Joint
  Conference on Artificial Intelligence, {IJCAI-17}}, pp.\  2627--2633, 2017.
\newblock \doi{10.24963/ijcai.2017/366}.
\newblock URL \url{https://doi.org/10.24963/ijcai.2017/366}.

\bibitem[Shao et~al.(2024)Shao, Wang, Zhu, Xu, Song, Bi, Zhang, Zhang, Li, Wu,
  and Guo]{shao2024deepseekmath}
Zhihong Shao, Peiyi Wang, Qihao Zhu, Runxin Xu, Junxiao Song, Xiao Bi, Haowei
  Zhang, Mingchuan Zhang, Y.~K. Li, Y.~Wu, and Daya Guo.
\newblock {DeepSeekMath}: Pushing the limits of mathematical reasoning in open
  language models.
\newblock \emph{arXiv preprint arXiv:2402.03300}, 2024.
\newblock URL \url{https://arxiv.org/abs/2402.03300}.

\bibitem[Shi et~al.(2025)Shi, Song, Zhang, Shi, Luo, Ao, Arian, and
  Seco]{shi2025alphaforge}
Hao Shi, Weili Song, Xinting Zhang, Jiahe Shi, Cuicui Luo, Xiang Ao, Hamid
  Arian, and Luis~Angel Seco.
\newblock {AlphaForge}: A framework to mine and dynamically combine formulaic
  alpha factors.
\newblock In \emph{Proceedings of the AAAI Conference on Artificial
  Intelligence}, volume~39, pp.\  12524--12532, 2025.
\newblock \doi{10.1609/aaai.v39i12.33365}.

\bibitem[Shi et~al.(2026{\natexlab{a}})Shi, Yan, Cai, and Lv]{hubble}
Runze Shi, Shengyu Yan, Yuecheng Cai, and Chengxi Lv.
\newblock Hubble: An llm-driven agentic framework for safe, diverse, and
  reproducible alpha factor discovery.
\newblock \emph{arXiv preprint arXiv:2604.09601}, 2026{\natexlab{a}}.
\newblock URL \url{https://arxiv.org/abs/2604.09601}.

\bibitem[Shi et~al.(2026{\natexlab{b}})Shi, Duan, and Li]{shi2026alphajungle}
Yu~Shi, Yitong Duan, and Jian Li.
\newblock Navigating the alpha jungle: An {LLM}-powered {MCTS} framework for
  formulaic alpha factor mining.
\newblock In \emph{Proceedings of the AAAI Conference on Artificial
  Intelligence}, volume~40, pp.\  997--1005, 2026{\natexlab{b}}.
\newblock \doi{10.1609/aaai.v40i2.37069}.

\bibitem[Tang et~al.(2025)Tang, Chen, Yang, Mai, Zheng, Wang, Chen, and
  Lin]{tang2025alphaagent}
Ziyi Tang, Zechuan Chen, Jiarui Yang, Jiayao Mai, Yongsen Zheng, Keze Wang,
  Jinrui Chen, and Liang Lin.
\newblock {AlphaAgent}: {LLM}-driven alpha mining with regularized exploration
  to counteract alpha decay.
\newblock In \emph{Proceedings of the 31st ACM SIGKDD Conference on Knowledge
  Discovery and Data Mining}, 2025.
\newblock \doi{10.1145/3711896.3736838}.
\newblock URL \url{https://arxiv.org/abs/2502.16789}.

\bibitem[Tang et~al.(2026)Tang, Yin, Chen, Chen, Zheng, Ye, Wang, and
  Lin]{tang2026alphaagentevo}
Ziyi Tang, Xuexiong Yin, Weixing Chen, Zechuan Chen, Yongsen Zheng, Wenxuan Ye,
  Keze Wang, and Liang Lin.
\newblock Alphaagentevo: Evolution-oriented alpha mining via self-evolving
  agentic reinforcement learning.
\newblock In \emph{The Fourteenth International Conference on Learning
  Representations}, 2026.
\newblock URL \url{https://openreview.net/forum?id=lNmZrawUMu}.

\bibitem[Vaswani et~al.(2017)Vaswani, Shazeer, Parmar, Uszkoreit, Jones, Gomez,
  Kaiser, and Polosukhin]{vaswani2017attention}
Ashish Vaswani, Noam Shazeer, Niki Parmar, Jakob Uszkoreit, Llion Jones,
  Aidan~N. Gomez, {\L}ukasz Kaiser, and Illia Polosukhin.
\newblock Attention is all you need.
\newblock In \emph{Advances in Neural Information Processing Systems}, 2017.
\newblock URL \url{https://arxiv.org/abs/1706.03762}.

\bibitem[Wang et~al.(2026{\natexlab{a}})Wang, Yin, Fu, Zhang, Zeng, and
  Fong]{alphamaster}
Haozengran Wang, Shuo Yin, Rong Fu, Mengting Zhang, Jiayi Zeng, and Simon~James
  Fong.
\newblock Alphamaster: Dual-chain feedback for scalable and diverse alpha
  factor discovery.
\newblock In \emph{Proceedings of the 32nd ACM SIGKDD Conference on Knowledge
  Discovery and Data Mining V. 2}, pp.\  4882--4893, 2026{\natexlab{a}}.
\newblock \doi{10.1145/3770855.3818120}.

\bibitem[Wang et~al.(2026{\natexlab{b}})Wang, Xu, Zhang, Huang, Sun, and
  Zhang]{factorminer}
Yanlong Wang, Jian Xu, Hongkang Zhang, Shao-Lun Huang, Danny~Dongning Sun, and
  Xiao-Ping Zhang.
\newblock Factorminer: A self-evolving agent with skills and experience memory
  for financial alpha discovery.
\newblock In \emph{Proceedings of the 32nd ACM SIGKDD Conference on Knowledge
  Discovery and Data Mining V. 2}, pp.\  12327--12338, 2026{\natexlab{b}}.

\bibitem[White(2000)]{white2000reality}
Halbert White.
\newblock A reality check for data snooping.
\newblock \emph{Econometrica}, 68\penalty0 (5):\penalty0 1097--1126, 2000.
\newblock \doi{10.1111/1468-0262.00152}.

\bibitem[Xia et~al.(2024)Xia, Ao, Li, Liu, Liu, Ye, and Chai]{xia2024cisthpan}
Hongjie Xia, Huijie Ao, Long Li, Yu~Liu, Sen Liu, Guangnan Ye, and Hongfeng
  Chai.
\newblock Ci-sthpan: Pre-trained attention network for stock selection with
  channel-independent spatio-temporal hypergraph.
\newblock In \emph{Proceedings of the AAAI conference on artificial
  intelligence}, volume~38, pp.\  9187--9195, 2024.

\bibitem[Xiao et~al.(2024)Xiao, Sun, Luo, and Wang]{xiao2024tradingagents}
Yijia Xiao, Edward Sun, Di~Luo, and Wei Wang.
\newblock {TradingAgents}: Multi-agents {LLM} financial trading framework.
\newblock \emph{arXiv preprint arXiv:2412.20138}, 2024.
\newblock URL \url{https://arxiv.org/abs/2412.20138}.

\bibitem[Xiong et~al.(2025)Xiong, Deng, Wang, Cao, Li, Yu, Peng, Lin, Smith,
  Liu, Huang, Ananiadou, and Xie]{xiong2025flagtrader}
Guojun Xiong, Zhiyang Deng, Keyi Wang, Yupeng Cao, Haohang Li, Yangyang Yu,
  Xueqing Peng, Mingquan Lin, Kaleb~E. Smith, Xiao-Yang Liu, Jimin Huang,
  Sophia Ananiadou, and Qianqian Xie.
\newblock {FLAG}-{TRADER}: Fusion {LLM}-agent with gradient-based reinforcement
  learning for financial trading.
\newblock In \emph{Findings of the Association for Computational Linguistics:
  ACL 2025}, pp.\  13921--13934, 2025.
\newblock URL \url{https://aclanthology.org/2025.findings-acl.716/}.

\bibitem[Yang et~al.(2023)Yang, Liu, and Wang]{yang2023fingpt}
Hongyang Yang, Xiao-Yang Liu, and Christina~Dan Wang.
\newblock {FinGPT}: Open-source financial large language models.
\newblock \emph{arXiv preprint arXiv:2306.06031}, 2023.
\newblock URL \url{https://arxiv.org/abs/2306.06031}.

\bibitem[Yang et~al.(2020)Yang, Liu, Zhou, Bian, and Liu]{yang2020qlib}
Xiao Yang, Weiqing Liu, Dong Zhou, Jiang Bian, and Tie-Yan Liu.
\newblock {Qlib}: An {AI}-oriented quantitative investment platform.
\newblock \emph{arXiv preprint arXiv:2009.11189}, 2020.
\newblock URL \url{https://arxiv.org/abs/2009.11189}.

\bibitem[Yi et~al.(2026)Yi, Yang, Jin, Li, and Li]{alphaschema}
Jingyang Yi, Jian Yang, Yifei Jin, Yuqi Li, and Jian Li.
\newblock Alphaschema: Exploring the space of trading semantics for llm-based
  alpha mining.
\newblock \emph{arXiv preprint arXiv:2607.26642}, 2026.
\newblock URL \url{https://arxiv.org/abs/2607.26642}.

\bibitem[Yu et~al.(2022)Yu, Velu, Vinitsky, Gao, Wang, Bayen, and
  Wu]{yu2022mappo}
Chao Yu, Akash Velu, Eugene Vinitsky, Jiaxuan Gao, Yu~Wang, Alexandre Bayen,
  and Yi~Wu.
\newblock The surprising effectiveness of {PPO} in cooperative multi-agent
  games.
\newblock In \emph{Advances in Neural Information Processing Systems},
  volume~35, 2022.
\newblock URL
  \url{https://proceedings.neurips.cc/paper_files/paper/2022/hash/9c1535a02f0ce079433344e14d910597-Abstract.html}.

\bibitem[Yu et~al.(2026)Yu, Zheng, Pan, Liu, Wang, and He]{alphamemo}
Hang Yu, Zifan Zheng, Jeff~Z Pan, Tongliang Liu, Zhiyong Wang, and Fengxiang
  He.
\newblock Alphamemo: Structured search-process memory for self-evolving alpha
  mining agents.
\newblock \emph{arXiv preprint arXiv:2606.20625}, 2026.
\newblock URL \url{https://arxiv.org/abs/2606.20625}.

\bibitem[Yu et~al.(2023)Yu, Xue, Ao, Pan, He, Tu, and He]{alphagen}
Shuo Yu, Hongyan Xue, Xiang Ao, Feiyang Pan, Jia He, Dandan Tu, and Qing He.
\newblock Generating synergistic formulaic alpha collections via reinforcement
  learning.
\newblock In \emph{Proceedings of the 29th ACM SIGKDD conference on knowledge
  discovery and data mining}, pp.\  5476--5486, 2023.

\bibitem[Zhang et~al.(2026)Zhang, Jia, Zhai, Xie, Duan, He, Yu, and
  Li]{zhang2026quantevolver}
Lingzhe Zhang, Tong Jia, Yunpeng Zhai, Zixuan Xie, Chiming Duan, Minghua He,
  Philip~S. Yu, and Ying Li.
\newblock From feedback loops to policy updates: Reinforcement fine-tuning for
  {LLM}-based alpha factor discovery.
\newblock \emph{arXiv preprint arXiv:2605.15412}, 2026.
\newblock URL \url{https://arxiv.org/abs/2605.15412}.

\end{thebibliography}
\bibliographystyle{iclr2027_conference}

\clearpage
\appendix
\section{Additional Related Work}
\label{app:additional-related-work}

\noindent\textbf{Traditional Alpha Mining.}
Machine learning estimates nonlinear return relations~\citep{gu2020empirical,chen2024deeplearning}, while latent-factor and graph models capture shared risks and stock interactions~\citep{duan2022factorvae,qian2024mdgnn,xia2024cisthpan,duan2025factorgcl}. Symbolic search and reinforcement learning instead construct factor collections~\citep{alphagen,shi2025alphaforge}. Selecting among many candidates also creates data-snooping and adaptive-selection risks~\citep{white2000reality,cawley2010overfitting,dwork2015adaptive}. These methods supply predictors and evaluation tools; our learning target is the policy that conducts factor research.

\noindent\textbf{LLM-Based Alpha Mining.}
Financial adaptation and reward-based training support prediction and trading~\citep{yang2023fingpt,xiong2025flagtrader,deng2026alphaquanter}, with neural-symbolic mining and evolutionary post-training extending learning to factor generation~\citep{fama_agent,tang2026alphaagentevo,zhang2026quantevolver}. Research workflows coordinate analysis, proposals, implementation, and feedback~\citep{xiao2024tradingagents,li2025rdagent,tang2025alphaagent,guo2026aqua,liu2026cognitive}; external memory reuses earlier successes and failures~\citep{factorminer,xalpha,alphamemo}. Automated scientific search~\citep{lu2024aiscientist,ning2026specialist} also faces evaluation leakage~\citep{luo2025pitfalls}, motivating separate search feedback and frozen final evaluation~\citep{ning2026molprop,ning2026materials}. We combine this separation with post-training of local planning and realization policies.

\noindent\textbf{Research Diversity.}
Expression regularization and tree search reduce formula redundancy~\citep{tang2025alphaagent,shi2026alphajungle}; family control and heterogeneous generators broaden factor pools~\citep{hubble,alphamaster}. Semantic plans expose the ideas behind implementations~\citep{alphaschema}. However, distinct outputs can concentrate on similar ideas~\citep{audranreiss2025ideation,chen2026diversitycollapse}. We therefore evaluate economic mechanisms and numerical signals together, and use complementary plans and varied research environments to construct post-training data.

\noindent\textbf{Joint Reinforcement Learning.}
Cooperative RL addresses credit assignment through counterfactual advantages and centralized training~\citep{foerster2018coma,yu2022mappo}. For LLM systems, MARFT models interdependent agent decisions~\citep{liao2025marft}, MAGRPO learns collaboration from shared rewards~\citep{liu2025magrpo}, and M-GRPO coordinates hierarchical planner--executor updates~\citep{hong2025mgrpo}. Our joint training specializes this direction to alpha mining: Planner credit compares research portfolios, Realizer credit compares implementations of the same plans, and their shared reward combines predictive quality with complementary prediction changes.

\section{Methodology Details}
\label{app:method-examples}

Table~\ref{tab:method-dataflow} follows one initial research round. The plan descriptions are shortened for presentation; the two factor computations in Figure~\ref{fig:spec-example} come from an executed API example for one plan's specifications.

\begin{table}[!t]
\centering\small
\caption{An example of the data passed between research modules. The symbols follow Section~\ref{sec:methodology}.}
\label{tab:method-dataflow}
\renewcommand{\arraystretch}{1.18}
\setlength{\tabcolsep}{7pt}
\begin{tabular}{@{}p{0.19\linewidth}p{0.75\linewidth}@{}}
\toprule
\rowcolor{MethodTint}\textbf{Object} & \textbf{Example and role} \\
\midrule
Environment $\mathcal E$ &
$\mathcal U$: historical CSI500 constituents; $y$: next open-to-open excess return; $\mathcal V$: daily and minute price--volume features; $\mathcal B$: 21 initial factors; $\mathcal O$: registered operators; $\mathcal G$: mechanisms feasible for these features and operators; $\eta$: inner time folds and later outer evaluation. \\
Initial state $s_1$ &
$(\mathcal E,\mathcal B,\varnothing)$: the environment, the initial factor set, and no earlier research feedback. \\
Plan portfolio $\mathcal P_1$ &
$p_{1,1}$: market residual continuation;\newline
$p_{1,2}$: late-session pressure reversal;\newline
$p_{1,3}$: volume--price confirmation;\newline
$p_{1,4}$: range compression. \\
Expanded $p_{1,4}$ &
\textbf{Event:} unusually compressed trading range.\newline
\textbf{Mechanism:} temporary balance may precede renewed price movement.\newline
\textbf{Horizon:} 1 day. \textbf{Condition:} distinguish quiet from fragile states.\newline
\textbf{Direction:} nonlinear. \textbf{Relation to base:} a new direction. \\
Retrieved cards $\mathcal K_{1,4}$ &
\textbf{Features:} \texttt{high\_low\_range} (daily relative range), \texttt{minute\_open\_5m\_range} (opening-window range).\newline
\textbf{Operators:} \texttt{ts\_rank} (trailing within-stock percentile rank), \texttt{at\_divide} (protected ratio).\newline
\textbf{Transforms:} \texttt{identity}, \texttt{at\_reverse} (negation). \\
Specifications $\mathcal Z_{1,4}$ &
Two explicit implementations of the range-compression plan, shown in Figure~\ref{fig:spec-example}. Other plans receive their own cards. \\
Evidence and next state &
Execution supplies numerical factors; screening and model evaluation produce $e_1$ and $u_1$. Analysis produces $a_1$. The updated factor set and history form $s_2=(\mathcal E,\mathcal F_1,h_1)$. \\
\bottomrule
\end{tabular}
\end{table}

\begin{figure}[!t]
\centering
\setlength{\fboxsep}{6pt}
\fcolorbox{DraftBlue}{MethodTint}{\begin{minipage}[t]{0.45\linewidth}
\small\textbf{Realization 1: historical compression}\par\medskip
\footnotesize\ttfamily\raggedright
plan\_id: p4\par
operator: ts\_rank\par
input\_features:\par
\quad [high\_low\_range]\par
input\_transforms:\par
\quad [at\_reverse]\par
window: 60\par
expected\_direction: nonlinear\par\medskip
\normalfont\footnotesize
\textbf{Rationale:} a high rank of the negated daily range identifies unusually compressed recent trading.
\end{minipage}}
\hfill
\fcolorbox{DraftBlue}{MethodTint}{\begin{minipage}[t]{0.45\linewidth}
\small\textbf{Realization 2: opening-range ratio}\par\medskip
\footnotesize\ttfamily\raggedright
plan\_id: p4\par
operator: at\_divide\par
input\_features:\par
\quad [minute\_open\_5m\_range,\par
\quad\phantom{[}high\_low\_range]\par
input\_transforms:\par
\quad [identity, identity]\par
expected\_direction: nonlinear\par\medskip
\normalfont\footnotesize
\textbf{Rationale:} the opening range relative to the daily range describes where intraday compression occurs.
\end{minipage}}
\caption{Two specification cards for the same plan. The actual response uses a \texttt{realizations} list containing \texttt{plan\_id} and its two \texttt{factor\_specs}. The first computation ranks a historical series; the second forms an intraday ratio. Unused parameters are omitted. Names, formulas, and data layers are assigned by the system, not the Realizer.}
\label{fig:spec-example}
\end{figure}

\textbf{Economic-mechanism catalogue.}
Table~\ref{tab:mechanism-catalogue} lists the full catalogue of 38 mechanisms in six themes. Each entry records an economic explanation, required feature and construction capabilities, and typical horizons. For a given environment, $\mathcal G$ retains the entries supported by $(\mathcal V,\mathcal O)$ and stays fixed throughout the loop. Which mechanisms have been explored and how they performed are recorded in $h_{r-1}$.

\begin{table}[!tbp]
\centering\small
\caption{The full economic-mechanism catalogue. The environment-specific $\mathcal G$ contains its feasible entries.}
\label{tab:mechanism-catalogue}
\renewcommand{\arraystretch}{1.12}
\setlength{\tabcolsep}{5pt}
\begin{tabular}{@{}p{0.20\linewidth}cp{0.68\linewidth}@{}}
\toprule
\rowcolor{MethodTint}\textbf{Theme} & \textbf{Count} & \textbf{Mechanisms} \\
\midrule
Trend and momentum & 5 & short horizon momentum; medium horizon momentum; trend strength efficiency; breakout continuation; return acceleration \\
Reversal & 6 & short horizon reversal; overreaction exhaustion; opening gap reversal; late session pressure reversal; VWAP dislocation reversal; range oscillation \\
Liquidity and volume & 6 & liquidity shock; illiquidity price impact; trading-activity surprise; volume price confirmation; volume price disagreement; liquidity dry up \\
Intraday structure & 6 & opening pressure persistence; late session persistence; intraday return asymmetry; intraday volume concentration; high low timing; intraday VWAP pressure \\
Volatility and range & 7 & realized volatility regime; range expansion; range compression; volatility of volatility; downside upside asymmetry; tail risk extremes; intraday distribution shape \\
Relative returns and crowding & 8 & market relative strength; market beta exposure; idiosyncratic volatility; market residual continuation; market residual reversal; cross sectional dispersion; co movement crowding; dependence regime \\
\bottomrule
\end{tabular}
\end{table}

\subsection{Screening and Data Selection Details}
\label{app:selection-rules}
\label{app:factor-screening-details}

\textbf{Factor metrics and acceptance.}
For a factor $f$, coverage is the proportion of eligible inner stock--date observations for which both $f$ and the target $y$ are available. IC, RIC, ICIR, and RICIR follow Appendix~\ref{app:metric-definitions}, with $f$ replacing the model prediction. Let $R_{\mathrm{LS}}(f)$ denote the annualized return of the top-quintile minus bottom-quintile portfolio ranked by $f$. For the one-day target, annualization compounds these daily long--short returns at 252 trading days per year. The factor quality score is
\begin{equation}
 s_{\mathrm{fac}}(f)=\frac{1}{5}\left(
 \frac{|\mathrm{RIC}(f)|}{0.05}+\frac{|\mathrm{RICIR}(f)|}{0.35}
 +\frac{|\mathrm{IC}(f)|}{0.03}+\frac{|\mathrm{ICIR}(f)|}{0.25}
 +\frac{|R_{\mathrm{LS}}(f)|}{0.30}\right).
 \label{eq:factor-quality-score}
\end{equation}
The absolute values allow a consistently negative signal to be useful to the downstream predictor. The predictor score $S_k$ uses the same scales on fold $k$ but keeps the signed metrics, since the fitted predictor determines prediction direction. If a factor metric is unavailable, the score averages its available components; empty or constant factors are rejected.

For factors $f$ and $g$, redundancy is measured by the mean absolute daily cross-sectional correlation, $\bar\rho(f,g)=|\mathcal I|^{-1}\sum_{t\in\mathcal I}|\operatorname{Corr}_i(f_{i,t},g_{i,t})|$, where $\mathcal I$ contains dates with enough valid paired observations. Candidates are processed in decreasing factor score, so a redundant candidate is rejected in favor of a stronger retained candidate. Table~\ref{tab:factor-data-thresholds} distinguishes research-loop acceptance from the stricter quality gate used to select Realizer training pairs.

\begin{table}[!t]
\centering\small
\caption{Thresholds for research factors and Realizer pair examples. Both factors in a training pair must pass the right-hand column and have a recorded pair correlation.}
\label{tab:factor-data-thresholds}
\renewcommand{\arraystretch}{1.14}
\begin{tabular}{@{}p{0.44\linewidth}cc@{}}
\toprule
\rowcolor{MethodTint}\textbf{Criterion} & \textbf{Research acceptance} & \textbf{Realizer pair selection} \\
\midrule
Factor--target coverage & $\geq0.20$ & $\geq0.80$ \\
Days with valid IC & $\geq2$ & $\geq120$ \\
Factor score $s_{\mathrm{fac}}$ & $\geq0.05$ & $\geq0.25$ \\
Absolute RIC & Included in score & $\geq0.01$ \\
Absolute RICIR & Included in score & $\geq0.05$ \\
Mean absolute correlation & $\leq0.90$ against references & $\leq0.80$ within the pair \\
\bottomrule
\end{tabular}
\end{table}

\textbf{Assignment and sampling.}
The three Planner groups are defined in Section~\ref{sec:training-data}. If an example meets several group definitions, the exporter assigns it in the order productive switch, model improvement, then new direction. This makes the groups disjoint. Each mechanism--implementation signature is retained at most twice; a complete portfolio remains the supervision target. 

\textbf{Realizer pairs and repair examples.}
\label{app:realizer-data-details}
Pair examples preserve the original plan and retrieved contents, and both implementations must also have passed research acceptance. Missing correlation evidence does not qualify as low correlation. Repair inputs contain a faulty response and its validation feedback; the target is the validated pair. The corruptions described in Section~\ref{sec:training-data} alter only the input response. All examples derived from the same source run, and examples sharing an identical prompt, remain in the same train/evaluation split.

\Needspace{10\baselineskip}
\subsection{Joint Reward and Optimization Details}
\label{app:joint-training-details}

\textbf{Supervised initialization.} Joint GRPO starts from the Planner and Realizer SFT policies trained with the complete data construction strategy in Section~\ref{sec:training-data}. Planner SFT uses the 50/30/20 mixture of model-improvement, productive-switch, and new-direction examples. Realizer SFT uses the selected factor pairs and repair examples. Each role has a separate LoRA adapter and a frozen copy of its SFT policy as the reference.

\phantomsection\label{app:joint-task-bank}
\textbf{Task states.} We combine four markets, three feature masks, and eight archive configurations to construct 96 fixed task states. Each state $s=(\mathcal E,h)$ specifies its market observations, return target, permitted features and operators, baseline factors $\mathcal B$, and research history. The baseline factor set is part of the environment $\mathcal E$. Historical plans summarized in $h$ provide the archive $\mathcal H_s$ used for diversity evaluation. Archive factors serve as redundancy references and are not added to the baseline model. We re-evaluate up to 16 archived plans using the same evaluator as new plans, excluding plans without effective factors or positive quality. Each of the 96 states is visited twice during training.

\phantomsection\label{app:joint-evaluator}
\textbf{LightGBM evaluation.} Each task uses four chronological inner segments: initial training, validation, additional fitting, and reward evaluation. Their nominal lengths are 252, 63, 126, and 63 trading days, with label-overlap purging at segment boundaries. LightGBM uses at most 200 trees and early stopping after 50 rounds without validation improvement. It is then refitted on the eligible observations preceding the reward segment. Baseline, portfolio, and individual-plan models are evaluated on the same reward rows.

Candidate factors first pass the research acceptance thresholds in Table~\ref{tab:factor-data-thresholds}, using baseline and archive factors as redundancy references. A plan's factor subset $\mathcal A_j$ is taken from the portfolio's accepted set $\mathcal A$, preserving column order. For any accepted factor set, $S_s$ applies the research loop's signed composite model score from Appendix~\ref{app:factor-screening-details} to the reward window. It combines IC, RIC, their information ratios, and long--short return using the same metric scales.

\phantomsection\label{app:joint-sampling-details}
\textbf{Nested sampling and token probabilities.} We use $J=4$, $M=4$, and $N=2$. For each state, the Planner samples four portfolios of four plans. The Realizer generates two independent responses per portfolio, each containing two factor specifications per plan and eight specifications in total. The two responses share the same plans and retrieved features and operators. A grammar derived from these resources restricts Realizer generation to permitted JSON fields and values. When computing probabilities under the sampling, current, and SFT-reference policies, all three distributions are renormalized over the same permitted tokens. This makes the ratios in Eq.~\ref{eq:joint-grpo} consistent with the sampling distribution. Planner generation is not grammar-constrained.

\phantomsection\label{app:noise-controls}
\textbf{Noise portfolios and plans.} We construct eight noise portfolios per market to estimate score gains that can arise from uninformative inputs after factor selection and model fitting. Each portfolio contains eight numerical factor columns. We independently draw standard Gaussian values for every stock--date entry of each column, then standardize each column cross-sectionally within each date. We divide the eight columns into four two-factor groups, which serve as noise plans for plan-level evaluation. Generation uses a fixed seed derived from the global seed and market identifier.

These control factors pass through the same factor admission, accepted-set selection, and LightGBM fitting procedure as candidate factors. We compute both portfolio and plan gains relative to the baseline using the same composite score. If all factors in a control portfolio or plan are rejected, we reuse the baseline predictions and assign a gain of zero. The controls estimate gains from independent Gaussian inputs; they do not preserve the temporal or cross-sectional dependence of observed factors.

\phantomsection\label{app:reward-calibration}
\textbf{Reward calibration.} Before policy optimization, we evaluate historical candidate actions and the fixed noise controls on inner training tasks. Calibration is shared across feature masks within each market. We set $\sigma_s$ to the larger of the 75th percentile of absolute historical candidate score gains and the 95th percentile of absolute noise-portfolio score gains. We set $\kappa_s$ to the nonnegative part of the 95th percentile of noise-plan score gains. Calibration must yield a positive, numerically usable $\sigma_s$; otherwise it fails and policy optimization does not start. All reported training runs passed this check. The calibrated scales and thresholds remain fixed during training.

\Needspace{10\baselineskip}
\phantomsection\label{app:prediction-direction-details}
\textbf{Prediction-change vectors.} The vector introduced in Section~\ref{sec:local-post-training} is defined explicitly as
{
\begin{small}
\begin{equation}
\begin{aligned}
  d_{j,i,t}&=\widehat y^{(j)}_{i,t}-\widehat y^{(0)}_{i,t},
  &\bar d_{j,t}&=\frac{1}{n_t}\sum_{i=1}^{n_t}d_{j,i,t},\\
  \widetilde b_j&=\operatorname{concat}_{t=1}^{T_s}
    \left[\frac{d_{j,i,t}-\bar d_{j,t}}{\sqrt{T_s n_t}}\right]_{i=1}^{n_t},
  &b_j&=\frac{\widetilde b_j}{\|\widetilde b_j\|_2}.
\end{aligned}
\label{eq:prediction-direction}
\end{equation}
\end{small}}
All vectors use a common ordering of the eligible reward-window stock--date observations. Concatenating the centered daily differences gives $\widetilde b_j\in\mathbb R^{\sum_t n_t}$. The daily weighting satisfies
{
\begin{equation}
 \|\widetilde b_j\|_2^2
 =\frac{1}{T_s}\sum_{t=1}^{T_s}
    \frac{1}{n_t}\sum_{i=1}^{n_t}(d_{j,i,t}-\bar d_{j,t})^2.
 \label{eq:prediction-day-weighting}
\end{equation}}
Thus, each date contributes its mean squared prediction change, rather than receiving more weight simply because it contains more stocks. The final normalization compares directions independently of their overall magnitude, while $q_j$ supplies the quality weight. A zero-norm difference receives $b_j=0$ and $q_j=0$. Archived plans use the same baseline, reward rows, weighting, and score calibration as new plans. Their qualities and vectors are computed before policy optimization and held fixed.

\phantomsection\label{app:diversity-reward-details}
\textbf{Incremental useful diversity.} The log determinant in Eq.~\ref{eq:useful-diversity} rewards quality-weighted coverage of prediction-change directions. High-quality plans receive more weight, and similar changes receive diminishing reward. Subtracting the archive term measures the additional coverage supplied by the new plans. With an empty archive and four plans of quality $q_j=1$, four orthogonal vectors give $D=1$, whereas four identical vectors give $D=\log(5)/(4\log2)\approx0.58$. Repeated directions can therefore receive positive but smaller rewards. If every $q_j=0$, then $D=0$. Existing archive coverage reduces the gain from similar new directions; sign-reversed vectors provide the same coverage as their originals.

\phantomsection\label{app:advantage-calibration}
\textbf{Advantage scales.} After reward calibration and archive construction, we draw fresh samples from the unchanged SFT policies. For each role, we compute the unscaled numerator of its conditional advantage in Eq.~\ref{eq:conditional-advantages}. We set $c_{\mathrm P}$ and $c_{\mathrm R}$ to the corresponding root mean squares, each with a lower bound of 0.1. These two scales remain fixed during policy optimization.

\phantomsection\label{app:invalid-rollouts}
\textbf{Invalid outputs and failed sampling groups.} RL evaluates each sampled response directly, without repair. A contract-invalid Realizer response receives $U=-1$. An invalid Planner response is assigned $U=-1$ for both downstream outcomes when computing Planner credit; the Realizer is not invoked, and no Realizer loss is computed for those outcomes. When a valid response yields no accepted factors, the evaluator reuses the baseline predictions and assigns $Q=D=U=0$. If data or execution infrastructure fails for any outcome, we skip the state's entire $M\times N=4\times2$ sampling group in both role updates.

\phantomsection\label{app:joint-optimization}
\textbf{Optimization and KL estimation.} Each role has its own optimizer. We use learning rate $10^{-6}$, clipping $\epsilon=0.2$, and KL weight $\beta_a=0.01$. Visiting each of the 96 task states twice gives 192 state visits, arranged into 48 updates with four states per update. Training sampling uses temperature 1 with reasoning disabled, up to 4096 response tokens, and a total context limit of 65,536 tokens. Losses are averaged over response tokens and then over responses for each role. Only the two LoRA adapters are updated; the base model, SFT references, old sampling probabilities, and advantages receive no gradients.

For a token sampled from the old policy, define $z_\ell^a=\log\pi_{\mathrm{SFT}}^a(v_\ell\mid c,v_{<\ell})-\log\pi_{\theta_a}(v_\ell\mid c,v_{<\ell})$. The sampled KL term is $\rho_\ell^a[\exp(z_\ell^a)-z_\ell^a-1]$, using the same token probabilities as the policy update. The SFT reference remains frozen throughout training.

\section{More Experimental Details}
\label{app:experimental-details}
\makeatletter\setlength{\@fptop}{0pt}\makeatother

\subsection{Data and Evaluation Protocol}
\label{app:data-and-evaluation}
\textbf{Stock universes and inputs.}
CSI300, CSI500, and CSI1000 represent large-, mid-, and small-cap Chinese stocks. We also evaluate the broader Shanghai/Shenzhen A-share universe. Eligible stocks follow historical daily index membership. Our source data comprise daily open, high, low, close, share volume, and traded amount, together with minute bars. We construct both the initial factor set $\mathcal B$ and the available feature set $\mathcal V$ from these observations.

As Table~\ref{tab:base-factor-definitions}  shows, the 21 base factors $\mathcal B$ describe daily returns, trading activity, candlestick shape, and relative prices. The feature set $\mathcal V$ additionally contains rolling return and volatility statistics, daily Alpha158 formulas, and intraday summaries of opening/closing returns, participation, ranges, and VWAP deviations. Feature masking removes a fixed fraction of $\mathcal V$ while retaining basic daily OHLC, volume, and amount; it leaves $\mathcal B$ unchanged. Table~\ref{tab:feature-operator-examples} illustrates $\mathcal V$ and the operator set $\mathcal O$.

\begin{table}[!t]\centering
\caption{All 21 initial factors in $\mathcal B$. For one stock and date, $o,h,l,c,v,a$ denote OHLC prices, share volume, and traded amount; $w=a/v$, $q=h-l$, $u=\max(o,c)$, and $d=\min(o,c)$. Zero denominators give missing values.}
\label{tab:base-factor-definitions}
\small\setlength{\tabcolsep}{5pt}\renewcommand{\arraystretch}{1.12}
\begin{tabular}{lclc}\toprule
Factor & Formula & Factor & Formula \\ \midrule
\texttt{ret\_1} & $c_t/c_{t-1}-1$ & \texttt{open\_close\_ret} & $c/o-1$ \\
\texttt{high\_low\_range} & $h/l-1$ & \texttt{close\_position} & $(c-l)/q$ \\
\texttt{vwap} & $w$ & \texttt{vwap\_deviation} & $c/w-1$ \\
\texttt{log\_turnover} & $\log(1+a)$ & \texttt{log\_volume} & $\log(1+v)$ \\
\texttt{KMID} & $(c-o)/o$ & \texttt{KLEN} & $q/o$ \\
\texttt{KMID2} & $(c-o)/q$ & \texttt{KUP} & $(h-u)/o$ \\
\texttt{KUP2} & $(h-u)/q$ & \texttt{KLOW} & $(d-l)/o$ \\
\texttt{KLOW2} & $(d-l)/q$ & \texttt{KSFT} & $(2c-h-l)/o$ \\
\texttt{KSFT2} & $(2c-h-l)/q$ & \texttt{OPEN0} & $o/c$ \\
\texttt{HIGH0} & $h/c$ & \texttt{LOW0} & $l/c$ \\
\texttt{VWAP0} & $w/c$ & & \\ \bottomrule
\end{tabular}\end{table}

\begin{table}[!t]\centering
\caption{Examples from available features $\mathcal V$ and permitted operators/transforms $\mathcal O$. Traded amount is monetary value; it is not the share-turnover rate.}
\label{tab:feature-operator-examples}
\small\setlength{\tabcolsep}{4pt}\renewcommand{\arraystretch}{1.12}
\begin{tabular}{p{.13\linewidth}p{.38\linewidth}p{.42\linewidth}}\toprule
Set & Examples & Interpretation \\ \midrule
$\mathcal V$, daily & \texttt{ret\_20}, \texttt{volatility\_20}, \texttt{RSV20} & Medium-horizon return, return variation, and position within a recent price range. \\
$\mathcal V$, minute & \texttt{last30\_ret}, \texttt{first\_half\_ret}, \texttt{close\_vwap\_deviation} & Closing/earlier-session moves and displacement from intraday VWAP. \\
$\mathcal O$, arithmetic & \texttt{at\_subtract}, \texttt{at\_divide}, \texttt{linear\_combo} & Differences, protected ratios, and weighted combinations. \\
$\mathcal O$, temporal & \texttt{ts\_mean}, \texttt{ts\_rank} & Trailing means and historical percentile ranks within each stock. \\
$\mathcal O$, transforms & \texttt{cs\_rank}, \texttt{cs\_zscore}, \texttt{at\_reverse} & Same-date ranking, standardization, and sign reversal. \\ \bottomrule
\end{tabular}\end{table}

\textbf{Prediction timing and eligible observations.}
All features use information available after close $t$. Let $o_{i,t}$ be stock $i$'s opening price and $b_t$ the benchmark return over the same future opening-to-opening interval. The target is
\begin{equation}
 y_{i,t}=\frac{o_{i,t+2}}{o_{i,t+1}}-1-b_t.
 \label{eq:excess-o2o-label}
\end{equation}
The A-share benchmark is the equal-weight return of eligible A-shares; the index universes use their corresponding index returns. We exclude Beijing-listed stocks, ST stocks, listings younger than 60 days, suspended/delisted observations, and observations rejected by the shared opening-price limit filters.

\textbf{Inner research and outer evaluation.}
The inner period is May 2022--December 2025. The main comparisons use one research fold with chronological training, validation, and search segments in a 7:1:2 ratio. Validation selects model settings, inner results enter the research loop. Boundary observations are purged according to the two-opening label lookahead. In the multi-fold variant, search segments do not overlap. A candidate factor block is retained only if it improves the weighted inner score and the required number of folds, as defined in Section~\ref{sec:agent-system}.

After research, factors and the model family are fixed. The last 15\% of inner observations selects final training settings, followed by fitting on all eligible inner observations. The resulting model is evaluated once on outer period. Standalone ML/DL predictors instead fit on the earlier 85\% of inner data and use the last 15\% for selection. The results for all agentic methods in this paper were obtained using all available features and an inner one-fold setup; feature masks and multi-fold settings were employed solely for the purpose of constructing diverse training data.

\subsection{Predictive and Portfolio Metrics}
\label{app:metric-definitions}
Let $\hat y_{i,t}$ be the predicted score and $y_{i,t}$ the future excess return of stock $i$ on signal date $t$. The information coefficient (IC) measures Pearson correlation, and rank IC (RIC) measures Spearman correlation:
\begin{equation}
 \mathrm{IC}_t=\operatorname{Corr}_i(\hat y_{i,t},y_{i,t}),\qquad
 \mathrm{RIC}_t=\operatorname{Corr}_i\bigl(\operatorname{rank}(\hat y_{i,t}),\operatorname{rank}(y_{i,t})\bigr).
\end{equation}
IC and RIC are date averages. ICIR and RICIR divide each average by its sample standard deviation across dates.

Portfolio metrics are annualized excess return (ARR) for the equally weighted top 20\% of predicted stocks, information ratio (IR), maximum drawdown (MDD), and Calmar ratio (CR). Let $L_t$ be the top-quintile portfolio, $r_{i,t}$ the realized stock return, and $b_t$ the benchmark return. Its daily excess return is $e_t=|L_t|^{-1}\sum_{i\in L_t}r_{i,t}-b_t$. With 252 trading days per year,
\begin{equation}
 \mathrm{ARR}=252\bar e,\qquad
 \mathrm{IR}=\sqrt{252}\frac{\bar e}{\operatorname{sd}(e_t)},\qquad
 W_0=1,\quad W_t=\prod_{s=1}^{t}(1+e_s),
\end{equation}
\begin{equation}
 \mathrm{MDD}=\max_t\left(1-\frac{W_t}{\max_{0\leq s\leq t}W_s}\right),\qquad
 \mathrm{CR}=\frac{\mathrm{ARR}}{\mathrm{MDD}}.
\end{equation}
MDD is a positive loss magnitude, so lower is better. ARR is arithmetic annualized excess return; $W_t$ is compounded excess wealth. Financial tables use this same long-only portfolio definition throughout. Reported returns are gross of trading costs.

\subsection{Baselines Implementation}
\label{app:baseline-reproduction}

\textbf{ML and DL baselines.}

\noindent\textbf{Ridge}~\citep{gu2020empirical}. A linear return predictor with $\ell_2$ regularization selected on the validation segment.
\textbf{LightGBM}~\citep{ke2017lightgbm}. Gradient-boosted decision trees model nonlinear interactions among the initial factors.
\textbf{MLP}~\citep{gu2020empirical}. A feed-forward network combines the current factor observations to predict returns.
\textbf{GRU}~\citep{cho2014gru}. Gated recurrent units summarize temporal dependencies over the lookback sequence.
\textbf{LSTM}~\citep{hochreiter1997lstm}. Memory cells and input, output, and forget gates model longer temporal dependencies.
\textbf{ALSTM}~\citep{alstm}. Attention weights the recurrent states to emphasize informative historical observations.
\textbf{TCN}~\citep{bai2018tcn}. Causal temporal convolutions combine observations at different lookback positions.
\textbf{Transformer}~\citep{vaswani2017attention}. Self-attention models dependencies across the lookback sequence.
\textbf{PatchTST}~\citep{nie2023patchtst}. Time-series patches form tokens for attention over local temporal patterns.
\textbf{iTransformer}~\citep{liu2024itransformer}. Feature histories form tokens, so attention models relations among input variables.
\textbf{MASTER}~\citep{li2024master}. Market-conditioned attention combines temporal information with interactions among stocks.
\textbf{StockMixer}~\citep{fan2024stockmixer}. Mixing layers combine feature, temporal, and stock-set information; shared stock-set parameters accommodate changing historical constituents.

Following~\citet{li2025rdagent,li2026autoscientistquant}, we give every ML/DL baselines the same base factors to isolate the prediction architecture. Sequence models use 20 trading days. Neural models use a 64-dimensional hidden representation, dropout 0.1, AdamW with learning rate $10^{-3}$ and weight decay $10^{-5}$, and at most 30 epochs with six-epoch early stopping. The training objective is mean squared error on daily standardized returns; gradient norms are clipped at one. Model-specific adaptations preserve patch/feature tokenization or stock interactions while adding a scalar-return prediction head. Neural results are metric-wise medians across three seeds. Ridge selects its regularization on validation; LightGBM uses at most 500 trees and early stopping after 50 rounds. 

MASTER derives its market context from the same permitted factor inputs. This and the dynamic stock-set mixer keep information availability fixed while supporting changing index constituents.

\textbf{Agentic baselines.}
The controlled implementations share the data, historical universes, causal execution boundary, inner feedback, and outer dates.

\noindent\textbf{RD-Agent(Q)}~\citep{li2025rdagent}. Research hypotheses are translated into factors and evaluated iteratively; the feedback guides the next proposal.
\textbf{AlphaAgent}~\citep{tang2025alphaagent}. Factor proposals are additionally checked for expression originality and complexity before model evaluation.
\textbf{QuantaAlpha}~\citep{han2026quantaalpha}. Independent original, mutated, and crossover trajectories explore alternative factor sets.
\textbf{AlphaSchema}~\citep{alphaschema}. A selector evaluates semantic research plans, each realized at two lookback periods.

For a matched generation budget, every method receives 160 candidate-factor slots. RD-Agent(Q) and AlphaAgent receive 20 batches of eight slots. QuantaAlpha receives six original, six mutated, and eight crossover tasks with eight slots each. AlphaSchema receives five batches of 16 plans, each with two realizations. AlphaDiverse uses 20 rounds of four plans with two specifications per plan. Following~\citet{li2025rdagent,tang2025alphaagent,han2026quantaalpha,alphaschema}, we report one complete mining run per agentic method--market configuration, it is the common protocol of agentic alpha factor mining methods because of the high API/GPU cost.

\clearpage
\begin{table}[H]\centering
\caption{Final SFT dataset sizes and token counts, including repair examples. Supervised tokens belong to the target assistant responses; sequence tokens are the sum of input and supervised tokens.}\label{tab:sft-training-statistics}
\small\setlength{\tabcolsep}{5pt}\renewcommand{\arraystretch}{1.10}
{\begin{tabular}{lrrrr}\toprule
Role & Examples & Input tokens & Supervised tokens & Sequence tokens \\ \midrule
Planner & 589 & 9,863,796 & 341,416 & 10,205,212 \\
Realizer & 1,006 & 2,654,667 & 402,463 & 3,057,130 \\
\bottomrule\end{tabular}}
\end{table}
\begin{table}[H]\centering
\caption{SFT examples by source configuration. The four dimension blocks summarize the same datasets and are not additive across blocks.}\label{tab:sft-source-breakdown}
\small\setlength{\tabcolsep}{6pt}\renewcommand{\arraystretch}{1.00}
{\begin{tabular}{llrrr}\toprule
Dimension & Group & Source loops & Planner examples & Realizer examples \\ \midrule
API backend & GPT-5.5 & 24 & 196 & 253 \\
 & Grok-4.6 & 24 & 206 & 359 \\
 & GLM-5.3 & 24 & 187 & 394 \\
\midrule
Market & CSI300 & 18 & 111 & 171 \\
 & CSI500 & 18 & 130 & 256 \\
 & CSI1000 & 18 & 205 & 311 \\
 & A-share & 18 & 143 & 268 \\
\midrule
Mask rate & 0\% & 24 & 195 & 362 \\
 & 50\% & 24 & 256 & 336 \\
 & 70\% & 24 & 138 & 308 \\
\midrule
Inner folds & $K=1$ & 36 & 346 & 534 \\
 & $K=3$ & 36 & 243 & 472 \\
\bottomrule\end{tabular}}
\end{table}

\subsection{Local Post-Training and Inference}
\label{app:models-hardware}
Planner and Realizer use Qwen3.8-27B with separate rank-64 LoRA adapters and scaling 128. SFT runs for one epoch and applies loss only to final assistant responses; related trajectories and repair derivatives remain in the same split. Joint GRPO starts from the two SFT policies with a frozen shared backbone and separate role references. Each state samples four Planner portfolios and two Realizer responses per portfolio. The learning rate is $10^{-6}$, clipping is 0.2, and the KL coefficient is 0.01. Quality and diversity receive weights 0.75 and 0.25. Training uses $8\times$H200 GPUs.

Training rollouts disable reasoning and use temperature 1 with a 4096-token output allowance. Evaluation uses temperature 0.1. The default local Planner, Realizer, and Analysis nodes enable reasoning with a 131,000-token context and a 32,768-token response allowance; only the formal final answer is parsed. Planner and Realizer use the post-trained policies, whereas Analysis uses the original Qwen3.8-27B. API research uses GPT-5.5, Grok-4.6, or GLM-5.3; response limits are 8192 for the three APIs. Inference may place the two research policies on one H200 and Analysis on another.

\textbf{Training-data sources and SFT statistics.} We construct the SFT datasets from 72 API research loops, covering three backends (GPT-5.5, Grok-4.6, and GLM-5.3), three random feature-mask rates (0\%, 50\%, and 70\%), four markets, and two inner-fold settings ($K=1$ and $K=3$). These are training-data source runs; held-out source runs used in the fixed-state experiment are separate. Applying the selection and repair procedures in Section~\ref{sec:training-data} yields 589 Planner and 1,006 Realizer training examples. Each example is a complete input--response pair. A Realizer target contains eight factor specifications for four plans, and the counts include repair examples.

Table~\ref{tab:sft-training-statistics} reports the final SFT sample and token counts. Table~\ref{tab:sft-source-breakdown} groups the same examples by source backend, market, mask rate, and fold setting. Each dimension sums to the same 72 source loops, 589 Planner examples, and 1,006 Realizer examples. The source loops are balanced across each dimension, while the final example counts reflect the role-specific selection and repair procedures.

\clearpage
\begingroup\setlength{\intextsep}{3pt}\captionsetup{skip=3pt}
\begin{table}[H]\centering
\caption{Actual GRPO response and token counts across training. These are generation counts, with input and output tokens reported separately.}\label{tab:grpo-generation-statistics}
\small\setlength{\tabcolsep}{5pt}\renewcommand{\arraystretch}{1.10}
{\begin{tabular}{lrrrr}\toprule
Role & Responses & Input tokens & Output tokens & Total tokens \\ \midrule
Planner & 768 & 6,329,344 & 517,551 & 6,846,895 \\
Realizer & 1,220 & 6,088,926 & 1,023,290 & 7,112,216 \\
\midrule
Total & 1,988 & 12,418,270 & 1,540,841 & 13,959,111 \\
\bottomrule\end{tabular}}
\end{table}
\begin{table}[H]\centering
\caption{Shared prompt card for independent catalogue annotation.}\label{tab:judge-card}
\fontsize{8}{9}\selectfont\renewcommand{\arraystretch}{1.00}
\begin{tabular}{p{.11\linewidth}p{.82\linewidth}}\toprule
Field & Instruction \\ \midrule
Input & Anonymous ID, executed formula, economic rationale, and the complete mechanism catalogue. All inputs are causal daily/minute price--volume features. \\
Task & Assign exactly one primary catalogue ID to the predictive economic mechanism. Identify the driver and event, not the factor name or operator syntax. \\
Evidence & Prioritize the executed formula when it conflicts with the rationale; flag the conflict. A normalization or volatility denominator alone is not a new mechanism. \\
Boundaries & Window and scale changes preserve the main mechanism. Distinguish persistence from reversal using direction and economic intent. Do not infer profitability or method identity. \\
Unknown & Use \emph{other} and define a genuinely new mechanism; use \emph{uninterpretable} when the available evidence is insufficient. Do not force a match. \\
Output & Exact anonymous ID, one primary ID, up to two diagnostic secondary IDs, short formula-grounded evidence, conflict flag, and a definition when using \emph{other}. \\ \bottomrule
\end{tabular}\end{table}
\endgroup

\textbf{GRPO generation statistics.} The 96 task states, each visited twice with four Planner samples, produce 768 Planner responses. As specified in Appendix~\ref{app:invalid-rollouts}, invalid Planner outputs do not invoke the Realizer. The recorded rollout collection therefore contains 1,220 Realizer responses rather than the maximum of 1,536. Table~\ref{tab:grpo-generation-statistics} reports actual generated responses and their input and output tokens. Skipped downstream calls produce no Realizer responses or tokens; the loss-masking rules follow Appendix~\ref{app:invalid-rollouts}.

\subsection{Diversity Metrics}
\label{app:diversity-protocol}
\textbf{Diversity Metrics.}
All coverage statistics use the same entries listed in Table~\ref{tab:mechanism-catalogue}. Let $\mathcal F$ be the new factors generated in a run, $\mathcal F^{\mathrm{ret}}\subseteq\mathcal F$ the new factors in its final inner-selected model, and $g(f)$ the primary catalogue entry for factor $f$. We compute $M=|\{g(f):f\in\mathcal F\}|$ and $U=|\{g(f):f\in\mathcal F^{\mathrm{ret}}\}|$.
Base factors are excluded. A mechanism contributes one unit to U if at least one of its factors survives model selection.

Signal clusters (C). For each pair of factors, we compute their correlation across stocks on each inner date and average the absolute correlations over dates. Complete-linkage clustering groups factors only when every pair in a group has similarity at least 0.9. C counts the resulting groups, so larger values indicate more numerically distinct signals. This calculation is performed within each run; sign-reversed copies belong to the same group. Constant signals are excluded, and insufficient overlap remains undefined rather than being treated as independence.

Pair corr. takes the median of the same mean absolute correlations, restricted to the two factors implementing each plan. We use it only for AlphaDiverse variants with paired implementations; it is undefined for Single Synthesis or other methods without paired plans.

\textbf{Mapping external methods.}
AlphaDiverse uses its recorded catalogue IDs to compute diversity metrics. For other agentic methods without these IDs, GPT-5.5 and Grok-4.6 independently label anonymized formulas and rationales using the same catalogue. We shuffle records separately for the two annotators and hide method names, original mechanism labels, retention decisions, and all performance results. For annotations where the two parties disagreed, we used a GPT-5.6-sol annotator to make the final decision. A formula outside the catalogue receives \emph{other} with a proposed definition; insufficient evidence receives \emph{uninterpretable}. Neither adds to M or U.

This mapping gives external methods a common reporting scale. Differences in their search spaces and factor representations can affect annotation, so the principal diversity comparison concerns changes within AlphaDiverse. Table~\ref{tab:judge-card} states the shared annotation instructions.

\clearpage
\begin{table}[H]
\centering
\caption{CSI500 predictive and portfolio performance. ARR/MDD are percentages.}
\label{tab:financial-csi500}
\footnotesize\setlength{\tabcolsep}{2.4pt}\renewcommand{\arraystretch}{1.06}
\begin{tabular}{llrrrrrrrr}
\toprule
\multirow{2}{*}{Category} & \multirow{2}{*}{Method} & \multicolumn{4}{c}{Predictive performance} & \multicolumn{4}{c}{Portfolio performance} \\
\cmidrule(lr){3-6}\cmidrule(lr){7-10}
& & IC$\uparrow$ & ICIR$\uparrow$ & RIC$\uparrow$ & RICIR$\uparrow$ & ARR$\uparrow$ & IR$\uparrow$ & MDD$\downarrow$ & CR$\uparrow$ \\
\midrule
 & Ridge & 0.0050 & 0.0232 & 0.0225 & 0.0976 & -3.25 & -0.156 & 17.83 & -0.182 \\
 & MLP & 0.0150 & 0.0734 & 0.0296 & 0.1347 & 15.01 & 0.810 & 11.42 & 1.342 \\
\multirow{-3}{*}{ML} & LightGBM & 0.0166 & 0.0910 & 0.0228 & 0.1146 & -6.74 & -0.432 & 15.65 & -0.431 \\
\midrule
 & GRU & 0.0201 & 0.1008 & \underline{0.0309} & 0.1567 & 16.40 & 0.941 & 10.55 & 1.555 \\
 & LSTM & 0.0164 & 0.0788 & 0.0295 & 0.1380 & \textbf{21.30} & \underline{1.233} & 8.40 & \textbf{2.537} \\
 & ALSTM & 0.0181 & 0.0942 & 0.0279 & 0.1436 & 18.25 & 1.121 & 12.11 & 1.507 \\
 & TCN & 0.0246 & 0.1682 & \underline{0.0309} & 0.1786 & \underline{18.88} & 1.153 & 14.77 & 1.382 \\
 & Transformer & 0.0091 & 0.0485 & 0.0211 & 0.1050 & 6.41 & 0.309 & 14.49 & 0.365 \\
 & PatchTST & 0.0130 & 0.0719 & -0.0117 & -0.0523 & 2.00 & 0.178 & 17.87 & 0.168 \\
 & iTransformer & 0.0145 & 0.0739 & 0.0178 & 0.1325 & 10.80 & 0.788 & 13.50 & 0.872 \\
 & MASTER & 0.0178 & 0.0716 & 0.0219 & 0.1117 & 11.46 & 0.648 & 10.90 & 1.051 \\
\multirow{-9}{*}{DL} & StockMixer & -0.0169 & -0.0681 & 0.0179 & 0.0680 & -14.67 & -0.577 & 29.25 & -0.501 \\
\midrule
 & RD-Agent(Q) & \underline{0.0261} & \underline{0.1737} & 0.0265 & \underline{0.1984} & 2.50 & 0.214 & \underline{8.26} & 0.303 \\
 & AlphaAgent & 0.0172 & 0.0949 & 0.0283 & 0.1642 & 1.53 & 0.087 & 12.68 & 0.121 \\
 & QuantaAlpha & 0.0250 & 0.1148 & \textbf{0.0368} & 0.1716 & 14.88 & 0.822 & 10.46 & 1.422 \\
 & AlphaSchema & 0.0202 & 0.1212 & 0.0195 & 0.1257 & 6.07 & 0.465 & 10.86 & 0.559 \\
\rowcolor{MethodTint}
\multirow{-5}{*}{Agentic} & AlphaDiverse & \textbf{0.0308} & \textbf{0.2130} & 0.0295 & \textbf{0.2296} & 16.19 & \textbf{1.245} & \textbf{7.30} & \underline{2.216}  \\
\bottomrule
\end{tabular}
\end{table}
\section{More Experimental Results}
\label{app:more-experimental-results}

\subsection{Additional Financial Results}
\label{app:additional-results}
To test whether the financial performance observed on CSI300 extends to other markets, we compare AlphaDiverse with the same ML, DL, and agentic baselines on CSI500, CSI1000, and A-share. All methods follow the data splits, input settings, and evaluation protocol in Appendix~\ref{app:experimental-details}. Tables~\ref{tab:financial-csi500}--\ref{tab:financial-ashare} report predictive quality and portfolio performance using the same eight metrics as Table~\ref{tab:main-results}.

As Tables~\ref{tab:financial-csi500}--\ref{tab:financial-ashare} show, AlphaDiverse achieves the highest IC, ICIR, RICIR, and IR and the lowest MDD in all three additional markets. The strongest ARR on CSI500 and CSI1000 and the strongest RIC on CSI500 and A-share are obtained by other methods.

\clearpage
\begin{table}[H]
\centering
\caption{CSI1000 predictive and portfolio performance, January--August 2026. ARR/MDD are percentages. Definitions follow Table~\ref{tab:main-results}.}
\label{tab:financial-csi1000}
\fontsize{8}{9}\selectfont\setlength{\tabcolsep}{2.4pt}\renewcommand{\arraystretch}{1.06}
\begin{tabular}{llrrrrrrrr}
\toprule
\multirow{2}{*}{Category} & \multirow{2}{*}{Method} & \multicolumn{4}{c}{Predictive performance} & \multicolumn{4}{c}{Portfolio performance} \\
\cmidrule(lr){3-6}\cmidrule(lr){7-10}
& & IC$\uparrow$ & ICIR$\uparrow$ & RIC$\uparrow$ & RICIR$\uparrow$ & ARR$\uparrow$ & IR$\uparrow$ & MDD$\downarrow$ & CR$\uparrow$ \\
\midrule
 & Ridge & 0.0015 & 0.0076 & 0.0219 & 0.0998 & -16.05 & -0.751 & 20.97 & -0.765 \\
 & MLP & 0.0134 & 0.0725 & 0.0230 & 0.1232 & 0.09 & 0.004 & 21.78 & 0.004 \\
\multirow{-3}{*}{ML} & LightGBM & 0.0059 & 0.0332 & 0.0226 & 0.1159 & -13.43 & -0.654 & 21.60 & -0.622 \\
\midrule
 & GRU & 0.0125 & 0.0660 & 0.0088 & 0.0557 & -3.24 & -0.276 & 11.78 & -0.228 \\
 & LSTM & -0.0011 & -0.0062 & 0.0083 & 0.0575 & -19.68 & -1.348 & 20.03 & -0.982 \\
 & ALSTM & -0.0011 & -0.0060 & 0.0175 & 0.0861 & -7.83 & -0.411 & 22.14 & -0.372 \\
 & TCN & 0.0046 & 0.0292 & 0.0159 & 0.0872 & -5.50 & -0.264 & 26.09 & -0.227 \\
 & Transformer & 0.0091 & 0.0551 & 0.0203 & 0.0881 & -10.79 & -0.522 & 27.22 & -0.419 \\
 & PatchTST & -0.0049 & -0.0197 & 0.0189 & 0.1027 & -9.23 & -0.518 & 28.29 & -0.400 \\
 & iTransformer & 0.0032 & 0.0182 & 0.0247 & 0.1478 & -0.93 & -0.079 & 21.46 & -0.059 \\
 & MASTER & 0.0129 & 0.0687 & 0.0290 & 0.1325 & \textbf{17.56} & \underline{1.118} & 12.15 & \underline{1.451} \\
\multirow{-9}{*}{DL} & StockMixer & 0.0179 & 0.1020 & \underline{0.0309} & 0.1572 & -8.09 & -0.351 & 28.56 & -0.267 \\
\midrule
 & RD-Agent(Q) & 0.0137 & 0.1004 & 0.0249 & 0.1669 & -2.96 & -0.193 & \underline{10.99} & -0.269 \\
 & AlphaAgent & 0.0108 & 0.0606 & 0.0266 & 0.1512 & -7.71 & -0.422 & 17.05 & -0.452 \\
 & QuantaAlpha & \underline{0.0217} & \underline{0.1792} & 0.0294 & \underline{0.2167} & 9.16 & 0.625 & 11.59 & 0.790 \\
 & AlphaSchema & 0.0130 & 0.0898 & 0.0239 & 0.1447 & -0.22 & -0.014 & 13.07 & -0.017 \\
\rowcolor{MethodTint}
\multirow{-5}{*}{Agentic} & AlphaDiverse & \textbf{0.0256} & \textbf{0.1827} & \textbf{0.0364} & \textbf{0.2530} & \underline{16.45} & \textbf{1.171} & \textbf{7.97} & \textbf{2.063} \\
\bottomrule
\end{tabular}
\end{table}
\begin{table}[H]
\centering
\caption{A-share predictive and portfolio performance, January--August 2026. ARR/MDD are percentages. Definitions follow Table~\ref{tab:main-results}.}
\label{tab:financial-ashare}
\fontsize{8}{9}\selectfont\setlength{\tabcolsep}{2.4pt}\renewcommand{\arraystretch}{1.06}
\begin{tabular}{llrrrrrrrr}
\toprule
\multirow{2}{*}{Category} & \multirow{2}{*}{Method} & \multicolumn{4}{c}{Predictive performance} & \multicolumn{4}{c}{Portfolio performance} \\
\cmidrule(lr){3-6}\cmidrule(lr){7-10}
& & IC$\uparrow$ & ICIR$\uparrow$ & RIC$\uparrow$ & RICIR$\uparrow$ & ARR$\uparrow$ & IR$\uparrow$ & MDD$\downarrow$ & CR$\uparrow$ \\
\midrule
 & Ridge & 0.0123 & 0.0713 & \textbf{0.0409} & 0.2105 & 7.51 & 0.649 & 8.18 & 0.918 \\
 & MLP & 0.0188 & 0.1146 & 0.0349 & 0.1948 & 9.90 & 0.986 & 7.73 & 1.249 \\
\multirow{-3}{*}{ML} & LightGBM & 0.0246 & 0.1781 & 0.0381 & 0.2019 & 2.98 & 0.275 & 11.19 & 0.266 \\
\midrule
 & GRU & 0.0147 & 0.1044 & 0.0172 & 0.1072 & -1.63 & -0.160 & 11.38 & -0.143 \\
 & LSTM & 0.0145 & 0.0991 & 0.0202 & 0.1251 & -4.25 & -0.355 & 14.33 & -0.297 \\
 & ALSTM & 0.0187 & 0.1366 & 0.0240 & 0.1912 & 6.78 & 0.780 & 9.68 & 0.700 \\
 & TCN & 0.0139 & 0.1047 & 0.0229 & 0.1384 & 0.47 & 0.057 & 8.64 & 0.069 \\
 & Transformer & 0.0232 & 0.2047 & \underline{0.0405} & 0.2272 & 8.71 & 0.745 & 11.11 & 0.784 \\
 & PatchTST & 0.0017 & 0.0090 & 0.0294 & 0.1520 & -0.95 & -0.070 & 17.26 & -0.055 \\
 & iTransformer & 0.0100 & 0.0729 & 0.0308 & 0.2225 & 7.37 & 0.901 & 7.12 & 0.833 \\
 & MASTER & 0.0098 & 0.0671 & 0.0270 & 0.1632 & 5.91 & 0.609 & 8.97 & 0.775 \\
\multirow{-9}{*}{DL} & StockMixer & 0.0245 & 0.1761 & 0.0389 & \underline{0.2330} & \underline{10.25} & 0.934 & 9.40 & 0.896 \\
\midrule
 & RD-Agent(Q) & \underline{0.0317} & \underline{0.2932} & 0.0329 & 0.2310 & 8.48 & 0.849 & 6.33 & 1.339 \\
 & AlphaAgent & 0.0292 & 0.2344 & 0.0321 & 0.1868 & 2.91 & 0.271 & 8.99 & 0.324 \\
 & QuantaAlpha & 0.0302 & 0.2682 & 0.0328 & 0.2156 & 9.68 & \underline{1.027} & \underline{6.28} & \underline{1.541} \\
 & AlphaSchema & 0.0236 & 0.1923 & 0.0311 & 0.2014 & 0.96 & 0.091 & 10.86 & 0.088 \\
\rowcolor{MethodTint}
\multirow{-5}{*}{Agentic} & AlphaDiverse & \textbf{0.0357} & \textbf{0.3280} & 0.0392 & \textbf{0.2866} & \textbf{19.04} & \textbf{2.029} & \textbf{4.77} & \textbf{3.990} \\
\bottomrule
\end{tabular}
\end{table}

On CSI1000, it also achieves the highest RIC of 0.0364 and CR of 2.063, with an ARR of 16.45\%. On A-share, it improves ARR to 19.04\% and IR to 2.029, while reducing MDD to 4.77\%. Overall, the results extend the main observation to different stock universes: the post-trained local workflow produces informative factors and strong risk-adjusted portfolio performance.

\clearpage
\begin{table}[H]\centering
\caption{Predictive performance, portfolio performance, and research diversity of API-based AlphaDiverse across four markets. ARR/MDD are percentages; M, U, and C denote mechanisms, useful mechanisms, and signal clusters, respectively.}
\label{tab:backend-comparison}
\footnotesize\setlength{\tabcolsep}{1.2pt}
\begin{tabular}{lrrrrrrrrrrr}\toprule
Backend & IC & ICIR & RIC & RICIR & ARR & IR & MDD & CR & M & U & C \\ \midrule
\multicolumn{12}{l}{\textit{CSI300}} \\
GPT-5.5 & 0.0388 & 0.1995 & 0.0398 & 0.2446 & 35.97 & 2.872 & 7.08 & 5.083 & 24 & 11 & 108 \\
Grok-4.6 & 0.0285 & 0.1540 & 0.0324 & 0.1986 & 21.53 & 1.503 & 7.70 & 2.796 & 30 & 17 & 116 \\
GLM-5.3 & 0.0373 & 0.2105 & 0.0372 & 0.2418 & 26.41 & 2.214 & 9.24 & 2.858 & 31 & 18 & 132 \\
\midrule
\multicolumn{12}{l}{\textit{CSI500}} \\
GPT-5.5 & 0.0286 & 0.1979 & 0.0347 & 0.2371 & 6.12 & 0.426 & 10.33 & 0.592 & 24 & 11 & 125 \\
Grok-4.6 & 0.0260 & 0.1681 & 0.0269 & 0.1845 & 3.72 & 0.290 & 12.60 & 0.295 & 33 & 14 & 119 \\
GLM-5.3 & 0.0184 & 0.1224 & 0.0235 & 0.1684 & -2.77 & -0.241 & 11.38 & -0.243 & 30 & 7 & 124 \\
\midrule
\multicolumn{12}{l}{\textit{CSI1000}} \\
GPT-5.5 & 0.0205 & 0.1381 & 0.0333 & 0.2258 & 1.99 & 0.143 & 10.84 & 0.184 & 25 & 19 & 127 \\
Grok-4.6 & 0.0231 & 0.1370 & 0.0387 & 0.2279 & 4.53 & 0.262 & 10.70 & 0.424 & 33 & 21 & 118 \\
GLM-5.3 & 0.0150 & 0.0943 & 0.0234 & 0.1469 & -13.11 & -0.792 & 17.40 & -0.753 & 35 & 28 & 133 \\
\midrule
\multicolumn{12}{l}{\textit{A-share}} \\
GPT-5.5 & 0.0330 & 0.2904 & 0.0391 & 0.2864 & 13.05 & 1.184 & 5.88 & 2.219 & 28 & 17 & 129 \\
Grok-4.6 & 0.0364 & 0.3364 & 0.0409 & 0.2953 & 15.14 & 1.581 & 5.10 & 2.967 & 33 & 23 & 116 \\
GLM-5.3 & 0.0333 & 0.3067 & 0.0405 & 0.2945 & 12.41 & 1.234 & 8.75 & 1.419 & 31 & 21 & 133 \\
\bottomrule\end{tabular}\par
\end{table}
\subsection{Results of API-based AlphaDiverse}
\label{app:api-results}
To evaluate the research workflow with different API backends, we instantiate AlphaDiverse with GPT-5.5, Grok-4.6, and GLM-5.3. Each backend supplies the Planner, Realizer, and Analysis nodes under the same 20-round budget and evaluation protocol. Table~\ref{tab:backend-comparison} reports predictive performance, portfolio performance, and research diversity across all four markets.

As Table~\ref{tab:backend-comparison} shows, GPT-5.5 achieves the highest ARR on CSI300 and CSI500, reaching 35.97\% and 6.12\%, respectively. It also leads IC and RIC in both markets. Grok-4.6 achieves the highest ARR and IR on CSI1000 and A-share, with ARR values of 4.53\% and 15.14\%. GLM-5.3 explores the most mechanisms on CSI300 and CSI1000 and produces the most signal clusters on CSI300, CSI1000, and A-share. Its 28 useful mechanisms on CSI1000 nevertheless accompany a negative outer ARR, showing that inner-retained coverage and outer returns capture different outcomes. These results show that AlphaDiverse supports diverse exploration with all three APIs, while predictive and portfolio performance depend on the backend and market.

\clearpage
\begin{table}[H]\centering
\caption{Research diversity across four markets. Best and second-best values within each market and metric are bold and underlined; ties receive the same mark.}\label{tab:diversity-all}
\small\setlength{\tabcolsep}{2.4pt}\renewcommand{\arraystretch}{1.12}
\begin{tabular}{lrrrrrrrrrrrr}\toprule
Method & \multicolumn{3}{c}{CSI300} & \multicolumn{3}{c}{CSI500} & \multicolumn{3}{c}{CSI1000} & \multicolumn{3}{c}{A-share} \\
\cmidrule(lr){2-4}\cmidrule(lr){5-7}\cmidrule(lr){8-10}\cmidrule(lr){11-13}
 & M$\uparrow$ & U$\uparrow$ & C$\uparrow$ & M$\uparrow$ & U$\uparrow$ & C$\uparrow$ & M$\uparrow$ & U$\uparrow$ & C$\uparrow$ & M$\uparrow$ & U$\uparrow$ & C$\uparrow$ \\ \midrule
RD-Agent(Q) & 18 & 3 & 67 & 25 & \underline{14} & 121 & 15 & 9 & 127 & 8 & 8 & 23 \\
AlphaAgent & 5 & 5 & 16 & 5 & 1 & 34 & 1 & 1 & 19 & 3 & 3 & 20 \\
QuantaAlpha & 11 & 3 & 58 & 22 & 4 & 64 & 21 & 4 & 63 & 13 & 5 & 59 \\
AlphaSchema & 18 & 2 & 96 & 19 & 7 & 96 & 19 & 10 & 99 & 18 & 6 & 94 \\
Single Synthesis & 19 & 8 & 122 & 20 & 10 & 99 & 21 & 11 & 100 & 22 & 12 & 101 \\
AlphaDiverse (GPT-5.5) & 24 & 11 & 108 & 24 & 11 & \underline{125} & 25 & 19 & 127 & 28 & 17 & 129 \\
AlphaDiverse (Grok-4.6) & 30 & 17 & 116 & 33 & \underline{14} & 119 & \underline{33} & 21 & 118 & \underline{33} & \underline{23} & 116 \\
AlphaDiverse (GLM-5.3) & 31 & \underline{18} & \textbf{132} & 30 & 7 & 124 & \textbf{35} & \underline{28} & \textbf{133} & 31 & 21 & \textbf{133} \\
AlphaDiverse (untrained) & \underline{33} & 11 & 121 & \underline{34} & 6 & 117 & 26 & 11 & 109 & 31 & 19 & 103 \\
AlphaDiverse (SFT-only) & 27 & 4 & 85 & 23 & 6 & 83 & 28 & 18 & 89 & 26 & 16 & 104 \\
\rowcolor{MethodTint}
AlphaDiverse & \textbf{34} & \textbf{23} & \underline{129} & \textbf{36} & \textbf{25} & \textbf{128} & \textbf{35} & \textbf{29} & \underline{130} & \textbf{36} & \textbf{28} & \underline{132} \\
\bottomrule\end{tabular}\end{table}
\subsection{Cross-Method Research Diversity}
\label{app:cross-method-diversity}
To further evaluate research-path diversity across markets, we extend the comparison in Table~\ref{tab:research-diversity} to all four stock universes. We include the agentic baselines, Single Synthesis, three API-based AlphaDiverse variants, untrained AlphaDiverse, and SFT-only AlphaDiverse. All methods receive 160 candidate-factor slots, and M, U, and C follow Appendix~\ref{app:diversity-protocol}.

As Table~\ref{tab:diversity-all} shows, AlphaDiverse achieves the highest useful-mechanism coverage in all four markets and the highest or joint-highest mechanism coverage. Relative to SFT-only AlphaDiverse, U increases from 4 to 23, 6 to 25, 18 to 29, and 16 to 28. We also observe the same distinction from Single Synthesis as in the main text: on CSI300, its 122 signal clusters approach AlphaDiverse's 129, but it retains only 8 mechanisms rather than 23. GLM-5.3 produces more signal clusters than the complete model on three markets, yet retains fewer useful mechanisms throughout. Together, these results support the benefit of organizing and post-training research around complementary plans that contribute to the retained model.

\clearpage
\begin{figure}[H]\centering
\includegraphics[width=\linewidth]{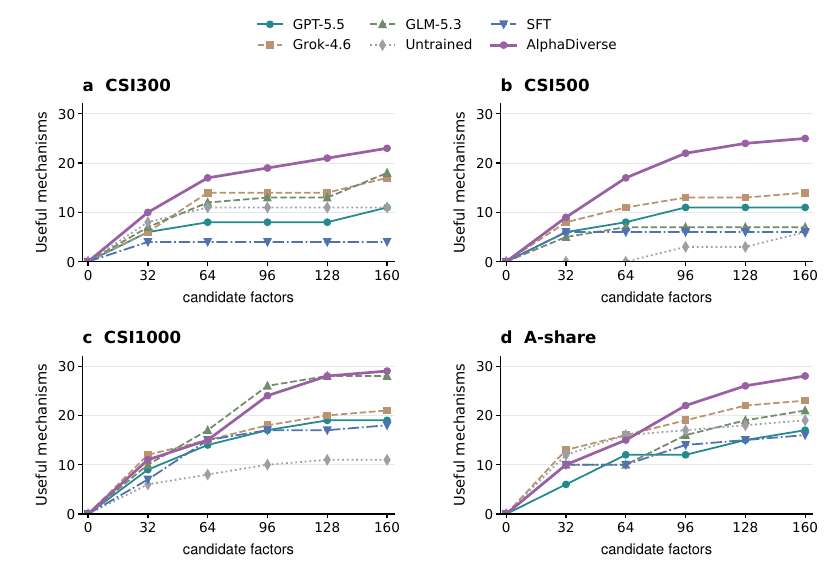}
\caption{Growth of useful-mechanism coverage with the number of candidate factors in four markets. Each curve follows one AlphaDiverse variant through its research loop.}\label{fig:coverage-all}
\end{figure}
\begin{table}[H]\centering
\caption{Model-retained mechanism coverage U at five candidate budgets.}\label{tab:coverage-all}
\scriptsize\setlength{\tabcolsep}{2.2pt}\renewcommand{\arraystretch}{1.08}
\resizebox{\linewidth}{!}{\begin{tabular}{lrrrrrrrrrrrrrrrrrrrr}\toprule
 & \multicolumn{20}{c}{Retained mechanisms $U\uparrow$} \\
Method & \multicolumn{5}{c}{CSI300} & \multicolumn{5}{c}{CSI500} & \multicolumn{5}{c}{CSI1000} & \multicolumn{5}{c}{A-share} \\
\cmidrule(lr){2-6}\cmidrule(lr){7-11}\cmidrule(lr){12-16}\cmidrule(lr){17-21}
 & 32 & 64 & 96 & 128 & 160 & 32 & 64 & 96 & 128 & 160 & 32 & 64 & 96 & 128 & 160 & 32 & 64 & 96 & 128 & 160 \\
\midrule
GPT-5.5 & 6 & 8 & 8 & 8 & 11 & 6 & 8 & 11 & 11 & 11 & 9 & 14 & 17 & 19 & 19 & 6 & 12 & 12 & 15 & 17 \\
Grok-4.6 & 6 & 14 & 14 & 14 & 17 & 8 & 11 & 13 & 13 & 14 & 12 & 15 & 18 & 20 & 21 & 13 & 16 & 19 & 22 & 23 \\
GLM-5.3 & 7 & 12 & 13 & 13 & 18 & 5 & 7 & 7 & 7 & 7 & 10 & 17 & 26 & 28 & 28 & 10 & 10 & 16 & 19 & 21 \\
Untrained & 8 & 11 & 11 & 11 & 11 & 0 & 0 & 3 & 3 & 6 & 6 & 8 & 10 & 11 & 11 & 12 & 16 & 17 & 18 & 19 \\
SFT & 4 & 4 & 4 & 4 & 4 & 6 & 6 & 6 & 6 & 6 & 7 & 15 & 17 & 17 & 18 & 10 & 10 & 14 & 15 & 16 \\
AlphaDiverse & 10 & 17 & 19 & 21 & 23 & 9 & 17 & 22 & 24 & 25 & 11 & 15 & 24 & 28 & 29 & 10 & 15 & 22 & 26 & 28 \\
\bottomrule\end{tabular}
}
\end{table}
\subsection{Four-Market Coverage and Transfer}
\label{app:coverage-results}
\textbf{Useful exploration over a research loop.} To test whether useful exploration persists as the candidate budget grows, we track U at 32, 64, 96, 128, and 160 candidate factors. Figure~\ref{fig:coverage-all} and Table~\ref{tab:coverage-all} compare API-based, untrained, SFT-only, and complete AlphaDiverse under the same settings as Figure~\ref{fig:research-diversity}.

As Figure~\ref{fig:coverage-all} and Table~\ref{tab:coverage-all} show, SFT-only AlphaDiverse reaches 4 and 6 useful mechanisms by 32 candidates on CSI300 and CSI500 and adds none thereafter. Complete AlphaDiverse continues to discover useful mechanisms, reaching 23 and 25 at 160 candidates. The same overall advantage appears on CSI1000 and A-share, where final coverage reaches 29 and 28, compared with 18 and 16 for SFT-only AlphaDiverse. Some API variants remain competitive, including GLM-5.3 with 28 useful mechanisms on CSI1000. Complete AlphaDiverse nevertheless achieves the highest final coverage in every market, supporting sustained discovery as research feedback accumulates.

\clearpage
\begin{figure}[H]\centering
\includegraphics[width=\linewidth]{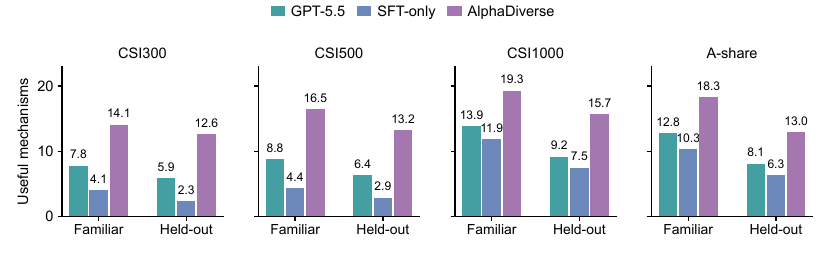}
\caption{Useful exploration from Familiar and Held-out source runs. For each market, 10 states are sampled from historical API runs that contributed to training-data construction and 10 from runs that did not. All three variants share these states. Bars show the union of useful mechanisms retained across eight independent single-round experiments from each state, averaged over the 10 states in each market and group. API-based AlphaDiverse uses GPT-5.5.}\label{fig:transfer-all}\label{tab:transfer-all}
\end{figure}
\textbf{Generalization at fixed research states.} To test whether useful exploration generalizes to held-out research histories, we compare API-based (GPT-5.5), SFT-only, and complete AlphaDiverse on shared research states. For each market, we pool historical runs generated by GPT-5.5, Grok-4.6, and GLM-5.3 and divide them according to whether they contributed to training-data construction. We randomly sample 10 states $s_r=(\mathcal E,\mathcal F_{r-1},h_{r-1})$ from contributing runs as the Familiar group and 10 from excluded runs as the Held-out group. Thus, group membership is defined by the source run, and all three variants receive the same 20 sampled states in each market. From every state, each variant performs eight independent single-round experiments, proposing four plans and two factor specifications per plan. Each experiment restores the saved environment, retained factor set, and research history before generation; its results do not alter the starting state of another experiment. Execution, factor screening, and model retention follow Section~\ref{sec:agent-system}, using inner data only. For each variant and state, we count the union of mechanisms represented by newly model-retained factors across the eight experiments. We then average these counts over the 10 states in each market and group.

As Figure~\ref{fig:transfer-all} shows, AlphaDiverse retains more useful alternatives than the API-based and SFT-only variants in both groups across all four markets. On CSI300, its Held-out coverage is 12.6, compared with 5.9 for GPT-5.5 and 2.3 for SFT-only AlphaDiverse. Consistent advantages appear on CSI500, CSI1000, and A-share, where AlphaDiverse reaches Held-out coverage of 13.2, 15.7, and 13.0, respectively. Coverage is lower in the Held-out group for every variant, while the complete model retains the strongest coverage in both groups. These results support useful exploration from research histories drawn from source runs excluded from training-data construction.

\clearpage
\begin{table}[H]\centering
\caption{Module ablations with GPT-5.5 across four markets. ARR/MDD are percentages.}\label{tab:workflow-complete}
\scriptsize\setlength{\tabcolsep}{2.2pt}\renewcommand{\arraystretch}{1.08}
\resizebox{\linewidth}{!}{\begin{tabular}{llrrrrrrrrrrrr}\toprule
\multirow{2}{*}{Market} & \multirow{2}{*}{Variant} & \multicolumn{8}{c}{Economic performance} & \multicolumn{4}{c}{Research diversity} \\\cmidrule(lr){3-10}\cmidrule(lr){11-14}
 &  & IC$\uparrow$ & ICIR$\uparrow$ & RIC$\uparrow$ & RICIR$\uparrow$ & ARR$\uparrow$ & IR$\uparrow$ & MDD$\downarrow$ & CR$\uparrow$ & M$\uparrow$ & U$\uparrow$ & C$\uparrow$ & Pair corr.$\downarrow$ \\
\midrule
CSI300 & - w/o complementarity & 0.0352 & 0.1901 & 0.0352 & \underline{0.2284} & \textbf{37.73} & \textbf{3.821} & \textbf{5.03} & \textbf{7.501} & 10 & 7 & 85 & 0.455 \\
 & - w/o retrieval & 0.0329 & 0.1692 & 0.0373 & 0.2166 & 28.39 & 2.256 & 11.28 & 2.517 & \textbf{25} & \underline{9} & \underline{116} & \underline{0.371} \\
 & - w/o memory & \underline{0.0370} & \underline{0.1939} & \underline{0.0383} & 0.2084 & 32.26 & 2.176 & \underline{5.29} & \underline{6.100} & 21 & \textbf{11} & 92 & 0.473 \\
 & Single Synthesis & 0.0319 & 0.1862 & 0.0316 & 0.2037 & 23.83 & 1.914 & 10.10 & 2.359 & 19 & 8 & \textbf{122} & --- \\
\rowcolor{MethodTint}
 & AlphaDiverse & \textbf{0.0388} & \textbf{0.1995} & \textbf{0.0398} & \textbf{0.2446} & \underline{35.97} & \underline{2.872} & 7.08 & 5.083 & \underline{24} & \textbf{11} & 108 & \textbf{0.322} \\
\midrule
CSI500 & - w/o complementarity & 0.0279 & \underline{0.1913} & 0.0293 & 0.2105 & 1.86 & 0.160 & 11.67 & 0.160 & 8 & 4 & 91 & 0.513 \\
 & - w/o retrieval & 0.0201 & 0.1266 & 0.0258 & 0.1640 & 1.11 & 0.100 & 12.99 & 0.085 & \underline{22} & \underline{10} & \underline{112} & \underline{0.484} \\
 & - w/o memory & 0.0216 & 0.1324 & 0.0292 & 0.1893 & 0.66 & 0.045 & 12.16 & 0.054 & 21 & 9 & 105 & \textbf{0.377} \\
 & Single Synthesis & \textbf{0.0289} & 0.1811 & \underline{0.0336} & \underline{0.2188} & \underline{5.71} & \underline{0.417} & \textbf{9.91} & \underline{0.576} & 20 & \underline{10} & 99 & --- \\
\rowcolor{MethodTint}
 & AlphaDiverse & \underline{0.0286} & \textbf{0.1979} & \textbf{0.0347} & \textbf{0.2371} & \textbf{6.12} & \textbf{0.426} & \underline{10.33} & \textbf{0.592} & \textbf{24} & \textbf{11} & \textbf{125} & 0.592 \\
\midrule
CSI1000 & - w/o complementarity & 0.0182 & \underline{0.1315} & 0.0275 & 0.2028 & -2.30 & -0.154 & \underline{11.98} & -0.192 & 7 & 6 & 92 & \textbf{0.274} \\
 & - w/o retrieval & 0.0172 & 0.1224 & \underline{0.0316} & \underline{0.2150} & \underline{-0.96} & \underline{-0.057} & 16.46 & \underline{-0.058} & \textbf{25} & 12 & \underline{113} & 0.425 \\
 & - w/o memory & 0.0164 & 0.1155 & 0.0228 & 0.1499 & -2.22 & -0.135 & 13.43 & -0.165 & \underline{22} & \underline{16} & 106 & 0.515 \\
 & Single Synthesis & \underline{0.0193} & 0.1303 & 0.0265 & 0.2115 & -5.03 & -0.351 & 13.47 & -0.373 & 21 & 11 & 100 & --- \\
\rowcolor{MethodTint}
 & AlphaDiverse & \textbf{0.0205} & \textbf{0.1381} & \textbf{0.0333} & \textbf{0.2258} & \textbf{1.99} & \textbf{0.143} & \textbf{10.84} & \textbf{0.184} & \textbf{25} & \textbf{19} & \textbf{127} & \underline{0.375} \\
\midrule
A-share & - w/o complementarity & \underline{0.0329} & 0.2811 & 0.0355 & \underline{0.2548} & \underline{11.20} & \underline{1.165} & 6.52 & \underline{1.719} & 9 & 7 & 93 & 0.632 \\
 & - w/o retrieval & 0.0286 & 0.2404 & \underline{0.0373} & 0.2507 & 8.85 & 0.862 & \textbf{5.63} & 1.572 & \underline{22} & \underline{14} & \underline{114} & 0.515 \\
 & - w/o memory & 0.0325 & 0.2792 & 0.0330 & 0.2219 & 8.12 & 0.753 & 9.31 & 0.873 & 21 & 10 & 107 & \underline{0.494} \\
 & Single Synthesis & 0.0322 & \underline{0.2886} & 0.0272 & 0.1935 & 6.03 & 0.597 & 10.81 & 0.558 & \underline{22} & 12 & 101 & --- \\
\rowcolor{MethodTint}
 & AlphaDiverse & \textbf{0.0330} & \textbf{0.2904} & \textbf{0.0391} & \textbf{0.2864} & \textbf{13.05} & \textbf{1.184} & \underline{5.88} & \textbf{2.219} & \textbf{28} & \textbf{17} & \textbf{129} & \textbf{0.456} \\
\bottomrule\end{tabular}
}
\end{table}
\subsection{More Ablation Results}
\label{app:full-ablations}\label{app:ablation-plan}
\label{app:ablation-interventions}
\textbf{Module ablations.} To test the contribution of each module in the research agent system, we evaluate the four settings in Section~\ref{sec:experiment-ablations} across all markets. \emph{w/o complementarity} removes the Planner's within-round mechanism/theme quotas and adjacent-round novelty requirement while retaining feasibility and quality checks. \emph{w/o retrieval} replaces plan-specific retrieved cards $\mathcal K_{r,j}$ with all currently legal feature and operator cards, leaving the plans and data access unchanged. \emph{w/o memory} removes success/failure/crowding memory and recent semantic history from the Planner, retaining current model metrics, retained-factor cards, and the previous-round mechanism quota. \emph{Single Synthesis} replaces the Planner--Realizer decomposition with one agent that directly emits eight factor specifications without plan-specific retrieval. All settings use GPT-5.5 and retain execution, factor screening, model validation, and Analysis under the same candidate budget.

As Table~\ref{tab:workflow-complete} shows, removing complementarity reduces mechanism coverage to 7--10 across the four markets, compared with 24--28 for the complete workflow. Although this variant achieves a higher ARR on CSI300, it explores fewer useful mechanisms in every market. Removing retrieval preserves broad mechanism coverage but reduces U and RIC throughout, showing the importance of relevant implementation resources. Removing memory also lowers RIC in all four markets, while Single Synthesis retains fewer useful mechanisms than the complete workflow. These results extend the main ablation findings across markets: complementary planning broadens research directions, retrieval supports useful implementations, and historical feedback improves subsequent choices.

\clearpage
\begin{table}[H]\centering
\caption{Post-training objectives: economic performance and research diversity.}\label{tab:objective-complete}
\scriptsize\setlength{\tabcolsep}{2.2pt}\renewcommand{\arraystretch}{1.08}
\resizebox{\linewidth}{!}{\begin{tabular}{llrrrrrrrrrrrr}\toprule
\multirow{2}{*}{Market} & \multirow{2}{*}{Variant} & \multicolumn{8}{c}{Economic performance} & \multicolumn{4}{c}{Research diversity} \\\cmidrule(lr){3-10}\cmidrule(lr){11-14}
 &  & IC$\uparrow$ & ICIR$\uparrow$ & RIC$\uparrow$ & RICIR$\uparrow$ & ARR$\uparrow$ & IR$\uparrow$ & MDD$\downarrow$ & CR$\uparrow$ & M$\uparrow$ & U$\uparrow$ & C$\uparrow$ & Pair corr.$\downarrow$ \\
\midrule
CSI300 & - w/o training & 0.0257 & 0.1432 & 0.0291 & 0.1797 & 22.49 & 1.720 & 10.05 & 2.238 & \underline{33} & 11 & \underline{121} & 0.413 \\
 & - w/ SFT & 0.0317 & 0.1880 & 0.0389 & 0.2366 & 24.77 & 2.144 & 6.62 & 3.745 & 27 & 4 & 85 & 0.511 \\
 & - w/ Planner only & 0.0344 & 0.1992 & 0.0397 & 0.2557 & 32.57 & 2.624 & 6.12 & 5.321 & 30 & 13 & 105 & 0.423 \\
 & - w/ Realizer only & 0.0354 & 0.2029 & 0.0400 & 0.2620 & 35.17 & 2.784 & 5.95 & 5.907 & 30 & 15 & 111 & 0.394 \\
 & - w/o diversity reward & \underline{0.0363} & \underline{0.2066} & \underline{0.0403} & \underline{0.2684} & \underline{37.77} & \underline{2.943} & \underline{5.79} & \underline{6.526} & 29 & \underline{18} & 118 & \underline{0.365} \\
\rowcolor{MethodTint}
 & AlphaDiverse & \textbf{0.0378} & \textbf{0.2128} & \textbf{0.0407} & \textbf{0.2790} & \textbf{42.10} & \textbf{3.210} & \textbf{5.51} & \textbf{7.641} & \textbf{34} & \textbf{23} & \textbf{129} & \textbf{0.316} \\
\midrule
CSI500 & - w/o training & 0.0183 & 0.1142 & 0.0167 & 0.1076 & 0.18 & 0.013 & 13.13 & 0.014 & \underline{34} & 6 & \underline{117} & 0.584 \\
 & - w/ SFT & 0.0225 & 0.1408 & 0.0212 & 0.1423 & -1.77 & -0.136 & 12.65 & -0.140 & 23 & 6 & 83 & 0.451 \\
 & - w/ Planner only & 0.0262 & 0.1733 & 0.0249 & 0.1816 & 6.31 & 0.485 & 10.24 & 0.616 & 29 & 15 & 103 & 0.359 \\
 & - w/ Realizer only & 0.0275 & 0.1841 & 0.0262 & 0.1947 & 9.01 & 0.693 & 9.44 & 0.954 & 26 & 17 & 110 & 0.329 \\
 & - w/o diversity reward & \underline{0.0287} & \underline{0.1950} & \underline{0.0274} & \underline{0.2078} & \underline{11.70} & \underline{0.900} & \underline{8.64} & \underline{1.355} & 25 & \underline{20} & \underline{117} & \underline{0.298} \\
\rowcolor{MethodTint}
 & AlphaDiverse & \textbf{0.0308} & \textbf{0.2130} & \textbf{0.0295} & \textbf{0.2296} & \textbf{16.19} & \textbf{1.245} & \textbf{7.30} & \textbf{2.216} & \textbf{36} & \textbf{25} & \textbf{128} & \textbf{0.247} \\
\midrule
CSI1000 & - w/o training & 0.0176 & 0.1275 & 0.0225 & 0.1439 & -5.86 & -0.419 & 12.66 & -0.463 & 26 & 11 & 109 & 0.538 \\
 & - w/ SFT & 0.0217 & 0.1290 & \textbf{0.0403} & 0.2420 & -1.09 & -0.064 & 14.26 & -0.076 & 28 & 18 & 89 & 0.556 \\
 & - w/ Planner only & 0.0235 & 0.1532 & \underline{0.0385} & 0.2470 & 6.80 & 0.492 & 11.43 & 0.595 & \underline{31} & 23 & 107 & 0.518 \\
 & - w/ Realizer only & 0.0240 & 0.1612 & 0.0380 & 0.2486 & 9.43 & 0.677 & 10.49 & 0.900 & \underline{31} & \underline{25} & 114 & 0.506 \\
 & - w/o diversity reward & \underline{0.0246} & \underline{0.1693} & 0.0374 & \underline{0.2502} & \underline{12.06} & \underline{0.862} & \underline{9.54} & \underline{1.264} & 30 & 24 & \underline{120} & \underline{0.493} \\
\rowcolor{MethodTint}
 & AlphaDiverse & \textbf{0.0256} & \textbf{0.1827} & 0.0364 & \textbf{0.2530} & \textbf{16.45} & \textbf{1.171} & \textbf{7.97} & \textbf{2.063} & \textbf{35} & \textbf{29} & \textbf{130} & \textbf{0.472} \\
\midrule
A-share & - w/o training & \underline{0.0355} & \underline{0.3120} & 0.0355 & 0.2481 & \underline{18.66} & \underline{1.963} & \underline{5.15} & \underline{3.622} & \underline{31} & 19 & 103 & 0.420 \\
 & - w/ SFT & 0.0301 & 0.2574 & 0.0301 & 0.1938 & 8.28 & 0.769 & 9.52 & 0.870 & 26 & 16 & 104 & 0.610 \\
 & - w/ Planner only & 0.0326 & 0.2892 & 0.0342 & 0.2356 & 13.12 & 1.336 & 7.38 & 1.777 & 30 & 21 & 117 & 0.466 \\
 & - w/ Realizer only & 0.0335 & 0.2998 & 0.0356 & 0.2495 & 14.74 & 1.525 & 6.67 & 2.209 & 29 & \underline{23} & 121 & 0.419 \\
 & - w/o diversity reward & 0.0343 & 0.3104 & \underline{0.0369} & \underline{0.2634} & 16.35 & 1.714 & 5.96 & 2.744 & 28 & \underline{23} & \underline{125} & \underline{0.371} \\
\rowcolor{MethodTint}
 & AlphaDiverse & \textbf{0.0357} & \textbf{0.3280} & \textbf{0.0392} & \textbf{0.2866} & \textbf{19.04} & \textbf{2.029} & \textbf{4.77} & \textbf{3.990} & \textbf{36} & \textbf{28} & \textbf{132} & \textbf{0.291} \\
\bottomrule\end{tabular}
}
\end{table}
\textbf{Post-training ablations.} To evaluate the post-training pipeline, we compare the same five settings as in Section~\ref{sec:experiment-ablations}. \emph{w/o training} uses the original pretrained Planner and Realizer, and \emph{w/ SFT} uses both supervised policies without GRPO. \emph{w/ Planner only} starts from the same SFT pair and updates $\theta_{\mathrm P}$ while freezing $\theta_{\mathrm R}$ at its SFT value. \emph{w/ Realizer only} reverses these roles. Both agents still generate outputs in every rollout, so the single-role settings isolate which policy is updated. \emph{w/o diversity reward} updates both policies but removes $D$ from the shared utility in Eq.~\ref{eq:useful-diversity}. Complete AlphaDiverse jointly trains both policies using predictive quality and useful diversity. The research architecture, evaluator, task inputs, and original Analysis model are shared across these settings.

As Table~\ref{tab:objective-complete} shows, SFT improves RIC on CSI300, CSI500, and CSI1000, but narrows useful coverage on CSI300 and A-share. Thus, supervised learning alone does not consistently preserve useful exploration. Updating either the Planner or the Realizer improves ARR over SFT in every market, while joint training without the diversity reward further improves ARR. Adding the diversity term then increases U and reduces pair correlation in all four markets, while also improving ARR, IR, MDD, and CR. On CSI300, U rises from 18 to 23 and pair correlation falls from 0.365 to 0.316. These consistent gains support aligning research plans with their implementations and explicitly rewarding complementary contributions beyond joint quality optimization.

\clearpage
\begin{table}[H]\centering
\caption{SFT-data ablations across four markets.}\label{tab:data-complete}
\scriptsize\setlength{\tabcolsep}{2.2pt}\renewcommand{\arraystretch}{1.08}
\resizebox{\linewidth}{!}{\begin{tabular}{llrrrrrrrrrrrr}\toprule
\multirow{2}{*}{Market} & \multirow{2}{*}{Variant} & \multicolumn{8}{c}{Economic performance} & \multicolumn{4}{c}{Research diversity} \\\cmidrule(lr){3-10}\cmidrule(lr){11-14}
 &  & IC$\uparrow$ & ICIR$\uparrow$ & RIC$\uparrow$ & RICIR$\uparrow$ & ARR$\uparrow$ & IR$\uparrow$ & MDD$\downarrow$ & CR$\uparrow$ & M$\uparrow$ & U$\uparrow$ & C$\uparrow$ & Pair corr.$\downarrow$ \\
\midrule
CSI300 & - w/ improvement only & 0.0281 & 0.1671 & 0.0371 & 0.2154 & 22.38 & 1.962 & 7.51 & 2.980 & 21 & 1 & 73 & 0.616 \\
 & - w/o balancing & \underline{0.0308} & \underline{0.1814} & \underline{0.0383} & \underline{0.2296} & \underline{23.95} & \underline{2.084} & \underline{6.96} & \underline{3.441} & \underline{25} & \underline{3} & \underline{81} & \underline{0.546} \\
 & - w/o paired supervision & 0.0273 & 0.1600 & 0.0365 & 0.2081 & 21.51 & 1.907 & 7.85 & 2.740 & 19 & 1 & 69 & 0.651 \\
\rowcolor{MethodTint}
 & full data SFT & \textbf{0.0317} & \textbf{0.1880} & \textbf{0.0389} & \textbf{0.2366} & \textbf{24.77} & \textbf{2.144} & \textbf{6.62} & \textbf{3.745} & \textbf{27} & \textbf{4} & \textbf{85} & \textbf{0.511} \\
\midrule
CSI500 & - w/ improvement only & 0.0196 & 0.1195 & \underline{0.0206} & \underline{0.1350} & -4.12 & -0.312 & 13.56 & -0.304 & 17 & 3 & 71 & 0.556 \\
 & - w/o balancing & \underline{0.0211} & \underline{0.1336} & 0.0194 & 0.1214 & \underline{-2.57} & \underline{-0.194} & \underline{12.94} & \underline{-0.198} & \underline{21} & \underline{5} & \underline{79} & \underline{0.486} \\
 & - w/o paired supervision & 0.0185 & 0.1122 & 0.0188 & 0.1143 & -4.93 & -0.376 & 13.86 & -0.356 & 15 & 2 & 67 & 0.591 \\
\rowcolor{MethodTint}
 & full data SFT & \textbf{0.0225} & \textbf{0.1408} & \textbf{0.0212} & \textbf{0.1423} & \textbf{-1.77} & \textbf{-0.136} & \textbf{12.65} & \textbf{-0.140} & \textbf{23} & \textbf{6} & \textbf{83} & \textbf{0.451} \\
\midrule
CSI1000 & - w/ improvement only & 0.0183 & 0.1084 & 0.0385 & 0.2216 & -3.43 & -0.244 & 15.18 & -0.226 & 22 & 15 & 77 & 0.661 \\
 & - w/o balancing & \underline{0.0203} & \underline{0.1223} & \underline{0.0397} & \underline{0.2357} & \underline{-1.84} & \underline{-0.127} & \underline{14.50} & \underline{-0.127} & \underline{26} & \underline{17} & \underline{85} & \underline{0.591} \\
 & - w/o paired supervision & 0.0171 & 0.1010 & 0.0379 & 0.2147 & -4.28 & -0.306 & 15.41 & -0.277 & 20 & 14 & 73 & 0.696 \\
\rowcolor{MethodTint}
 & full data SFT & \textbf{0.0217} & \textbf{0.1290} & \textbf{0.0403} & \textbf{0.2420} & \textbf{-1.09} & \textbf{-0.064} & \textbf{14.26} & \textbf{-0.076} & \textbf{28} & \textbf{18} & \textbf{89} & \textbf{0.556} \\
\midrule
A-share & - w/ improvement only & 0.0273 & 0.2368 & 0.0283 & 0.1721 & 5.88 & 0.580 & 10.41 & 0.564 & 20 & 13 & 92 & 0.715 \\
 & - w/o balancing & 0.0267 & 0.2295 & \underline{0.0295} & \underline{0.1868} & \underline{7.45} & \underline{0.701} & \underline{9.86} & \underline{0.756} & \underline{24} & \underline{15} & \underline{100} & \underline{0.645} \\
 & - w/o paired supervision & \underline{0.0296} & \underline{0.2508} & 0.0277 & 0.1650 & 5.01 & 0.527 & 10.75 & 0.466 & 18 & 12 & 88 & 0.750 \\
\rowcolor{MethodTint}
 & full data SFT & \textbf{0.0301} & \textbf{0.2574} & \textbf{0.0301} & \textbf{0.1938} & \textbf{8.28} & \textbf{0.769} & \textbf{9.52} & \textbf{0.870} & \textbf{26} & \textbf{16} & \textbf{104} & \textbf{0.610} \\
\bottomrule\end{tabular}
}
\end{table}
\textbf{Data ablations.} To evaluate the data selection strategy, we compare \emph{w/ improvement only}, \emph{w/o balancing}, \emph{w/o paired supervision}, and \emph{full data SFT}. All settings are evaluated immediately after SFT with matched training-token budgets and no GRPO. \emph{w/ improvement only} restricts Planner examples to rounds with $u_r=1$, excluding productive-switch and new-direction examples without model improvement. \emph{w/o balancing} retains the eligible examples and three-group mixture but removes within-group balancing across environments, API backends, and research stages. \emph{w/o paired supervision} removes pair-aware quality/correlation selection and pair-specific repair augmentation for the Realizer, while preserving the two-specification response format and inference-time checks. \emph{full data SFT} uses the complete data selection strategy. All other training settings remain unchanged.

As Table~\ref{tab:data-complete} shows, improvement-only selection reduces useful coverage in every market, supporting the inclusion of productive switches and new directions. Removing balancing also lowers U throughout. Removing paired supervision increases pair correlation and reduces U in every market; on CSI300, correlation rises from 0.511 to 0.651 and U falls from 4 to 1. Full data SFT achieves the highest U and lowest pair correlation across markets. These results support varied Planner decisions and distinct Realizer implementations.

\clearpage
\begin{table}[H]\centering
\caption{Factor screening on CSI300. Best and second-best values are bold and underlined, respectively.}\label{tab:quality-funnel}
\small\setlength{\tabcolsep}{7pt}\renewcommand{\arraystretch}{1.12}
{\begin{tabular}{lrrrrr}\toprule
Method & Exec.$\uparrow$ & Acc.$\uparrow$ & Ret.$\uparrow$ & Forms$\uparrow$ & $|\mathrm{RIC}|\uparrow$ \\ \midrule
Untrained & \underline{151} & \underline{112} & 17 & \underline{146} & \underline{0.0110} \\
SFT & \textbf{160} & 78 & 8 & 132 & 0.0108 \\
GPT-5.5 & \textbf{160} & 104 & \underline{35} & 144 & 0.0099 \\
\rowcolor{MethodTint}
AlphaDiverse & \textbf{160} & \textbf{118} & \textbf{56} & \textbf{150} & \textbf{0.0140} \\
\bottomrule\end{tabular}}
\end{table}
\begin{table}[H]\centering
\caption{Recorded token usage and local GPU allocation per CSI300 research run.}\label{tab:deployment-tokens}
\small\setlength{\tabcolsep}{5pt}\renewcommand{\arraystretch}{1.10}
{\begin{tabular}{lrrr}\toprule
Deployment & Input tokens & Output tokens & H200 GPU-hours \\ \midrule
GPT-5.5 & 991,350 & 77,537 & 0.000 \\
Grok-4.6 & 1,189,848 & 531,985 & 0.000 \\
GLM-5.3 & 1,259,744 & 97,275 & 0.000 \\
\rowcolor{MethodTint}
AlphaDiverse (local) & 999,386 & 183,497 & 1.436 \\
\bottomrule\end{tabular}}
\end{table}
\subsection{Factor Quality}
\label{app:factor-quality}
To examine how generated factors contribute to the final predictive model, we compare execution (Exec.), factor acceptance (Acc.), and model retention (Ret.) on CSI300. Table~\ref{tab:quality-funnel} also reports the number of distinct formulas (Forms) and factor-level predictive quality.

AlphaDiverse retains 56 factors, compared with 35 for GPT-5.5 and 8 for SFT-only AlphaDiverse. The untrained model produces more accepted factors than GPT-5.5 but retains fewer in the predictive model. Individually acceptable signals can therefore offer limited additional value when combined with existing factors. This result supports rewarding downstream contributions to the factor set in Section~\ref{sec:local-post-training}.

\subsection{Deployment Cost Details}
\label{app:deployment-details}
Table~\ref{tab:deployment-cost} measures the inference resources for one complete CSI300 research run with all features, one inner fold, and 20 rounds of four plans with two factor specifications per plan. The accounting covers Planner, Realizer, and Analysis. In the local configuration, all three nodes use Qwen3.8-27B with reasoning enabled. Planner and Realizer use their jointly GRPO-trained policies, while Analysis uses the original weights. Table~\ref{tab:deployment-tokens} records token usage and locally allocated H200 GPU-hours. Output counts include recorded reasoning tokens. GPU-hours equal the number of allocated GPUs multiplied by job duration.

\end{document}